%% file: main.tex
\documentclass[lettersize,journal]{IEEEtran}
\usepackage{amsmath,amsfonts}
\usepackage{algorithmic}
\usepackage{algorithm}
\usepackage{array}
\usepackage[caption=false,font=normalsize,labelfont=sf,textfont=sf]{subfig}
\usepackage{textcomp}
\usepackage{stfloats}
\usepackage{url}
\usepackage{verbatim}
\usepackage{graphicx}
\usepackage{hyperref}
\usepackage{cite}
\usepackage{makecell}
\usepackage[most]{tcolorbox}
 
\newcommand{\revise}[1]{{\color{black}#1}} 

\newcommand{\revision}[1]{{\color{black}#1}} 

\begin{document}
\bstctlcite{IEEEexample:BSTcontrol}  % forces references to use et al. after 6 authors, according to the IEEE reference style guide (https://tex.stackexchange.com/questions/164017/limiting-the-number-of-authors-in-the-references-with-ieeetran)
\title{Power from Potential: A Survey of Electrostatic Actuators for Haptics}

\author{Ahad M. Rauf$^*$,~\IEEEmembership{Member, IEEE}, Ran Zhou$^*$,~\IEEEmembership{Member, IEEE}, Eric Acome,~\IEEEmembership{Member, IEEE}, Madeline Balaam,~\IEEEmembership{Member, IEEE}, Sean Follmer,~\IEEEmembership{Member, IEEE}, Teng Han,~\IEEEmembership{Member, IEEE}, Craig Shultz,~\IEEEmembership{Member, IEEE}, Daniel Leithinger,~\IEEEmembership{Member, IEEE}
\thanks{$^*$ Equal contribution authors.}
\thanks{Ahad M. Rauf and Sean Follmer are with the Department of Mechanical Engineering at Stanford University, Stanford, California, United States of America. E-mail: ahadrauf@alumni.stanford.edu, sfollmer@stanford.edu}
\thanks{Ran Zhou and Madeline Balaam are with the Division of Media Technology and Interaction Design at the KTH Royal Institute of Technology, Stockholm, Sweden. E-mail: ranzhou@kth.se, balaam@kth.se}
\thanks{Eric Acome is with Artimus Robotics, Boulder, Colorado, United States of America. E-mail: eric@artimusrobotics.com}
\thanks{Teng Han is with the Institute of Software, Chinese Academy of Science, Beijing, China. E-mail: hanteng@iscas.ac.cn}
\thanks{Craig Shultz is with the Department of Electrical and Computer Engineering, University of Illinois at Urbana-Champaign, Champaign, Illinois, United States of America. He is also with Fluid Reality, Inc., Chicago, Illinois, United States of America. E-mail: shultz88@illinois.edu}
\thanks{Daniel Leithinger is with the Department of Design Tech at Cornell University, Ithaca, New York, United States of America. E-mail: dl2265@cornell.edu}
}

% \author{IEEE Publication Technology,~\IEEEmembership{Staff,~IEEE,}}



\maketitle

\begin{abstract}

% As haptic interfaces move closer to the body and integrate more seamlessly into everyday environments, there is a growing demand for soft, thin, silent, and energy efficient actuators. However, conventional haptic actuators, typically based on mechanical motors and temperature-responsive polymers operating at low voltages and high currents, often suffer from bulky form factors or poor energy efficiency. High-Voltage Electrostatic Actuators (HVEAs), which generate force by applying an electric field to localized charge concentrations using high voltages and ultra-low currents, are an emerging haptic actuator class that offers responsive, silent, and low-power actuation within compliant, lightweight, and customizable form factors. They deliver feedback across diverse haptic modalities, support integrated self-sensing, and can be readily scaled up through monolithic manufacturing techniques to produce high spatial resolution actuator arrays. Although HVEAs have received increasing interest in the haptics community, barriers remain due to limited commercial availability and a lack of interdisciplinary knowledge exchange. This paper presents a comprehensive review of HVEAs for haptics, focusing on electrostatic switchable adhesives, dielectric elastomer actuators, soft electrohydraulic actuators, and electro-kinetic pumps. It then analyzes each actuator's haptic capabilities and form factors, identifies key challenges and design considerations, and envisions opportunities for how these programmable materials can unlock novel haptic interactions.

As haptic interfaces integrate more seamlessly into wearables and everyday environments, they increasingly require actuators that are soft, thin, silent, and energy efficient. However, conventional motors and temperature-responsive polymers often struggle to deliver these properties due to their bulky form factors and high power consumption. High-Voltage Electrostatic Actuators (HVEAs), which generate force by applying an electric field to localized charge concentrations using high voltages and ultra-low currents, have recently emerged as a compelling alternative due to their fast, silent, and low-power operation within highly customizable and compliant form factors.

\revise{This paper presents a focused review of HVEAs for haptics, examining four major classes: electrostatic switchable adhesives, dielectric elastomer actuators, soft electrohydraulic actuators, and electrokinetic pumps. For each class, we describe their mechanisms that enable haptic output; characterize their bandwidths, force densities, and spatial scalability; and evaluate their versatility for rendering cutaneous and kinesthetic feedback across wearable and world-grounded interfaces. Through this cross-technology analysis, we identify common design constraints and emerging strategies for improving ergonomics, streamlining fabrication, and integrating self-sensing. We conclude by outlining where HVEAs are uniquely positioned to advance haptic interaction and highlighting key research directions needed to translate these technologies into practical systems.}
% Through this cross-technology analysis, we identify common design constraints, including charge management, dielectric reliability, and high-voltage integration, as well as emerging strategies such as monolithic array fabrication, embedded self-sensing, and hybrid actuation architectures.}

\end{abstract}


\input{1_introduction}
\input{2_mechanisms_for_electrostatic_actuators}

\input{3_integrating_electrostatic_actuators_into_haptic_systems}
\input{4_adapting_traditional_design_workflows_for_electrostatic_actuators}

\input{5_conclusions_and_future_directions}

\section*{Acknowledgments}
The authors thank Joe Mullenbach and Yanjun Chen for their help in organizing the UIST 2023 workshop that inspired this review \cite{Leithinger_Zhou_Acome_Rauf_Han_Shultz_Mullenbach_2023}. They also thank Jad Mahmoud Halabi and Panče Naumov for their insights in choosing comparison metrics for the Ashby plots in Fig. \ref{fig:lit_review_cutaneous_feedback}.

Ahad Rauf and Sean Follmer's work on this project is supported by the National Science Foundation award grant no. 2142782. Ran Zhou and Madeline Balaam's work on this project is funded by the European Union (ERC, Intimate Touch, 101043637). Views and opinions expressed are however those of the author(s) only and do not necessarily reflect those of the European Union or the European Research Council. Neither the European Union nor the granting authority can be held responsible for them.

\bibliographystyle{IEEEtran}
\bibliography{bibliography}

\vfill

\end{document}

%% file: 1_introduction.tex
\section{Introduction}
% \IEEEPARstart{O}{ver} the past two decades, haptic technology has evolved significantly, shifting from rigid to soft materials, from bulky enclosures to thin films, and from single-point actuation to high spatial- and temporal-resolution arrays. These advancements have greatly improved ergonomics and touch expressiveness \cite{Zhu_Biswas_Dinulescu_Kastor_Hawkes_Visell_2022, Yin_Hinchet_Shea_Majidi_2021}. 
\IEEEPARstart{A}{s} haptic interfaces move closer to the body, becoming wearable, on-skin, or seamlessly integrated into everyday environments, there is a growing demand for actuators that are compact, lightweight, silent, and fast-responding \cite{Zhu_Biswas_Dinulescu_Kastor_Hawkes_Visell_2022, Yin_Hinchet_Shea_Majidi_2021}. These requirements can be challenging to meet with conventional haptic actuators such as DC motors, solenoids, and pneumatic pumps, which are typically bulky; vibration motors, which are noisy; and shape memory alloys, which have slow response times due to cooling delays. Beyond their mechanical drawbacks, these actuators typically rely on resistive or inductive operating principles that demand low voltages ($<$24 V) but high continuous currents ($>$10 mA), resulting in significant resistive efficiency losses. 
% These conventional haptic actuators also typically rely on resistive and inductive operating principles, which require low voltages $<$24 V but large, continuous current draws of $>$10 mA, resulting in significant resistive efficiency losses. 

\begin{figure}
    \includegraphics[width=\linewidth]{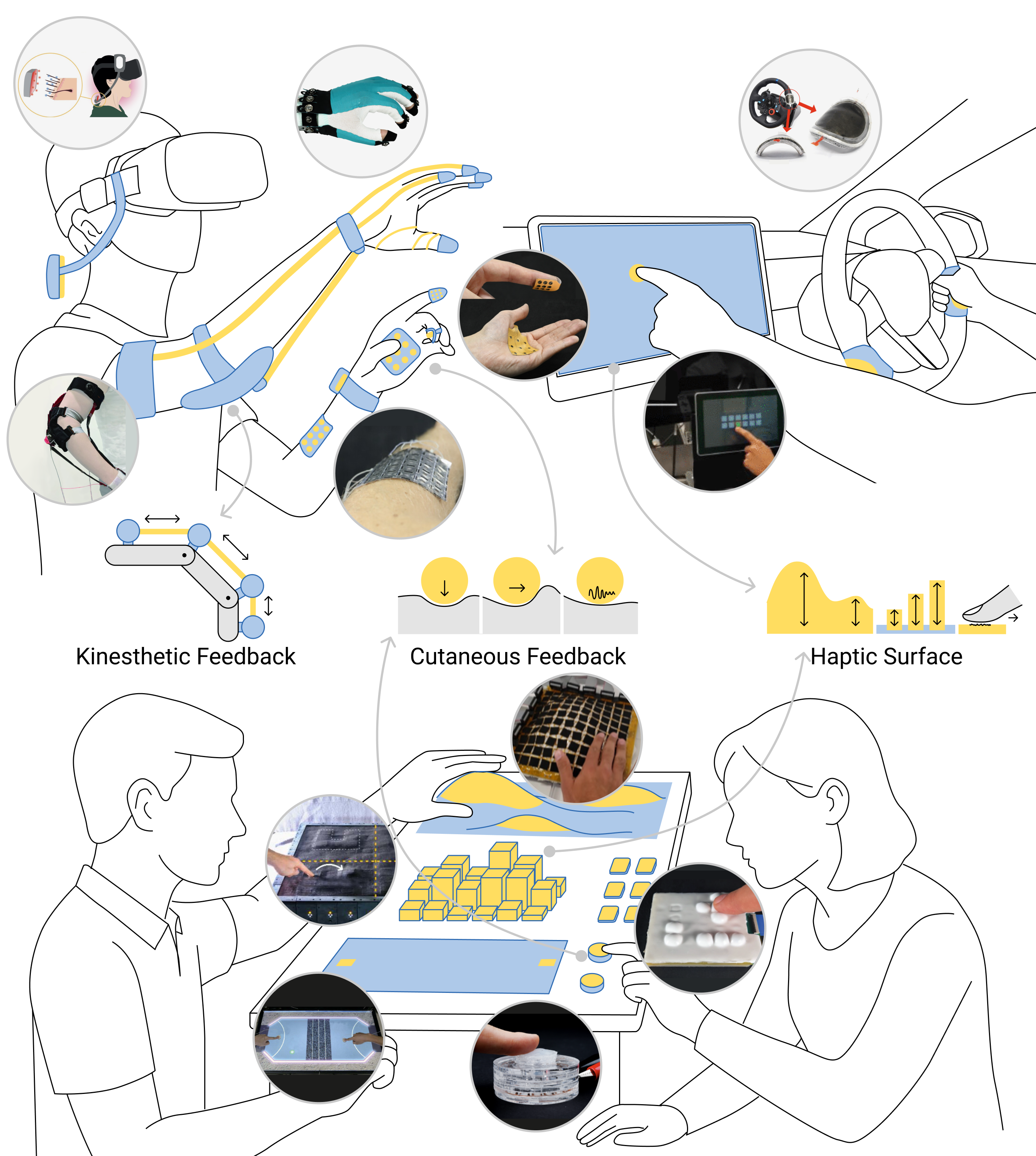}
    \caption{Example applications of HVEA haptic interfaces. Wearable devices are shown in the top left, providing cutaneous and kinesthetic feedback to render immersive virtual reality experiences. An in-car scenario is shown in the top right, using low-profile vibrotactile HVEAs in the steering wheel and touchscreen to communicate information without distracting drivers. A tabletop application is illustrated at the bottom, in which users can actively leverage large HVEA arrays to communicate dynamic spatio-temporal information. Starting from the top, images in bubbles were drawn from \cite{Iriarte_Ezcurdia_Elizondo_Irisarri_Hemmerling_Ortiz_Marzo_2024} (\copyright 2024 IEEE), \cite{Hinchet_Shea_2022} (\copyright 2022 The Authors. Advanced Intelligent Systems published by Wiley-VCH GmbH), \cite{Kim_Nam_Kim_Kyung_2021} (\copyright 2021 IEEE),\cite{Youn_Jang_Hwang_Pei_Yun_Kyung_2025} (\copyright 2025 The Authors), \cite{Ramachandran_Schilling_Wu_Floreano_2021} (\copyright 2021 The Authors. Advanced Intelligent Systems published by Wiley-VCH GmbH), \cite{sanchez2024cutaneous} (\copyright 2024 The Authors. Advanced Science published by Wiley‐VCH GmbH), \cite{Breitschaft_Pastukhov_Carbon_2022} (\copyright 2022 IEEE), \cite{Rauf_Bernardo_Follmer_2023} (\copyright 2023 IEEE), \cite{Johnson_Naris_Sundaram_Volchko_Ly_Mitchell_Acome_Kellaris_Keplinger_Correll_2023} (\copyright 2023 The Authors), \cite{Shultz_Harrison_2023} (\copyright 2023 The Authors), and \cite{Nakamura_Yamamoto_2016} (\copyright 2016 IEEE), \cite{fujii2021layerpump} (\copyright 2021 The Authors).}
    % Application scenarios of current HVEA haptic interfaces. Wearable devices (from top): images in bubbles were drawn from \cite{Iriarte_Ezcurdia_Elizondo_Irisarri_Hemmerling_Ortiz_Marzo_2024} (\copyright 2024 IEEE), \cite{Hinchet_Shea_2022} (\copyright 2022 The Authors. Advanced Intelligent Systems published by Wiley-VCH GmbH), \cite{Youn_Jang_Hwang_Pei_Yun_Kyung_2025} (\copyright 2025 The Authors), \cite{Ramachandran_Schilling_Wu_Floreano_2021} (\copyright 2021 The Authors. Advanced Intelligent Systems published by Wiley-VCH GmbH), \cite{sanchez2024cutaneous} (\copyright 2024 The Authors. Advanced Science published by Wiley‐VCH GmbH). In-car context (from top): \cite{Kim_Nam_Kim_Kyung_2021} (\copyright 2021 IEEE), \cite{Breitschaft_Pastukhov_Carbon_2022} (\copyright 2022 IEEE). Tabletop displays (from top): \cite{Rauf_Bernardo_Follmer_2023} (\copyright 2023 IEEE), \cite{Johnson_Naris_Sundaram_Volchko_Ly_Mitchell_Acome_Kellaris_Keplinger_Correll_2023} (\copyright 2023 The Authors), \cite{Shultz_Harrison_2023} (\copyright 2023 The Authors), \cite{Nakamura_Yamamoto_2016} (\copyright 2016 IEEE), \cite{fujii2021layerpump} (\copyright 2021 The Authors)
    \label{fig:tease figure}
\end{figure}

Recent advances in material science have resulted in a new class of technologies we refer to as High-Voltage Electrostatic Actuators (HVEAs)\revision{, which operate using high voltages ($>$ 100 V) and low currents ($<$ 1 mA). These include electrostatic switchable adhesives, dielectric elastomer actuators, soft electrohydraulic actuators, and electro-kinetic pumps, and example HVEA haptic applications are illustrated in Fig. \ref{fig:tease figure}.} HVEAs are built from lightweight and flexible materials, making them compact, near-silent, and fast-responding, and their form factor can be readily customized to deliver versatile cutaneous and kinesthetic haptic feedback. Operating at high voltages ($>$100 V) and low currents ($<$1 mA), HVEAs make use of capacitive mechanisms, which are orders of magnitude more energy dense than typical resistive or inductive mechanisms \cite{Mahmoud_Halabi_Ahmed_Sofela_Naumov_2021}. This enables HVEAs to achieve high force densities at much higher operating efficiencies than conventional haptic actuators. Their simple, customizable layer stacks integrate easily with diverse materials and form factors and support scalable production of high-spatial-resolution haptic interfaces. In addition, the capacitive property of HVEAs enables sensing capabilities to be integrated directly through circuitry, without requiring any changes to the mechanical actuator design. These attributes give HVEAs the potential to enable more ergonomic, higher-resolution, and more energy-efficient haptic interfaces that can advance a wide range of applications, such as mixed reality, teleoperation, healthcare, and assistive technologies.

% \revise{Recent advances in material science have resulted in a new class of technologies we refer to as High-Voltage Electrostatic Actuators (HVEAs), including but not limited to electrostatic switchable adhesives, dielectric elastomer actuators, soft electrohydraulic actuators, and electro-kinetic pumps.  HVEAs are built from lightweight and flexible materials, making them compact, near-silent, and fast-responding, and their form factor can be readily customized to deliver versatile cutaneous and kinesthetic haptic feedback. Operating at high voltages ($>$100 V) and low currents ($<$1 mA), HVEAs make use of capacitive mechanisms, which are orders of magnitude more energy dense than typical resistive or inductive mechanisms \cite{Mahmoud_Halabi_Ahmed_Sofela_Naumov_2021}. This enables HVEAs to achieve high force densities at much higher operating efficiencies than conventional haptic actuators. Their simple, customizable layer stacks integrate easily with diverse materials and form factors and support scalable production of high-spatial-resolution haptic interfaces. In addition, the capacitive property of HVEAs enables sensing capabilities to be integrated directly through circuitry, without requiring any changes to the mechanical actuator design.} These attributes give HVEAs the potential to enable more ergonomic, higher-resolution, and more energy-efficient haptic interfaces that can advance a wide range of applications, such as mixed reality, teleoperation, healthcare, and assistive technologies.
% \cite{May_2014} = good reference on energy density

\begin{figure}[t]
    \includegraphics[width=\linewidth]{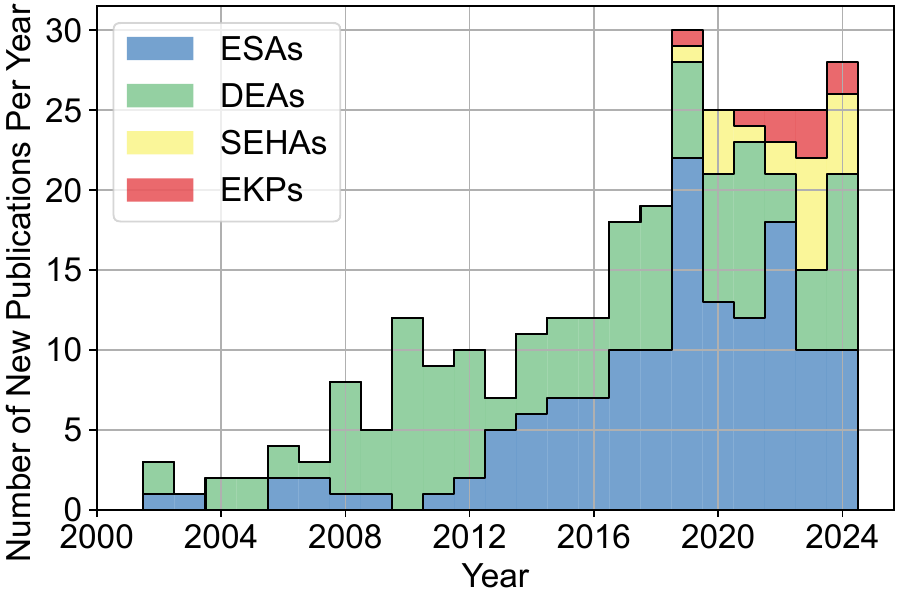}
    \caption{The number of new papers published each year that integrate HVEAs into haptic interfaces, aggregated across all HVEA classes in our review.}
    \label{fig:lit_review_num_papers_by_year}
\end{figure}

HVEAs have received increasing interest in the haptics and Human-Computer Interaction (HCI) communities, and as shown in Fig. \ref{fig:lit_review_num_papers_by_year} the number of papers published per year integrating HVEAs into haptic user interfaces has increased steadily over time. However, HVEAs still can have large barriers to entry because of their limited commercial availability and the interdisciplinary expertise required to design, prototype, and control them. Foundational research on these actuators is often published in robotics and material science venues, which provide useful references for their materials, operating principles, and performance capabilities \cite{Ankit_Ho_Nirmal_Kulkarni_Accoto_Mathews_2022, Perera_Liyanapathirana_Gargiulo_Gunawardana_2024, Rothemund_Kellaris_Mitchell_Acome_Keplinger_2021a, Vermes_Czigany_2020}. However, haptic and wearable applications involve unique design, ergonomics, and perceptual constraints are not discussed by these studies. Although several prior haptics review papers have mentioned HVEAs briefly as a part of larger discussions on surface haptics \cite{Basdogan_Giraud_Levesque_Choi_2020} and wearables for virtual and augmented reality \cite{Yin_Hinchet_Shea_Majidi_2021, Bai_Li_Shepherd_2021}, a systematic review of HVEA capabilities and design within the context of haptics has yet to be conducted. This survey aims to bridge this literature gap by synthesizing knowledge across domains, making HVEAs approachable for haptics researchers and designers. 
% Costes_Danieau_Argelaguet_Guillotel_Lecuyer_2020 (touchscreen-based surface haptics)

Inspired by the haptics taxonomy in Culbertson et al. \cite{culbertson2018haptics}, we conducted a systematic PRISMA literature review\cite{Page_McKenzie_Bossuyt_Boutron_Hoffmann_Mulrow_Shamseer_Tetzlaff_Akl_Brennan_2021} of 303 papers integrating HVEAs into haptic user interfaces.
We categorized prior work primarily by haptic modality and further mapped it across form factors, technologies, and application domains. \revise{Through this synthesis, we provide a comprehensive overview of how HVEAs have been designed, fabricated, and integrated into haptic interfaces, highlighting their potential to enable more ergonomic, expressive, and seamlessly integrated haptic systems. Building on these takeaways, we identify key opportunities for which HVEAs are uniquely positioned to advance haptic interaction, and we highlight future research directions needed to better translate these technologies into practical systems.} 

%% file: 2_mechanisms_for_electrostatic_actuators.tex
\section{Overview of High-Voltage Electrostatic Actuators} \label{sec:mechanisms_for_electrostatic_actuators}

% produce force from the Maxwell stress achieved
\revision{HVEAs} work by applying an electric field through a dielectric medium to produce a force on localized charge concentrations. They can be further categorized by the dielectric between their electrodes, and we identify HVEAs as: (a) electrostatic switchable adhesives (ESAs) when the dielectric is a vacuum or gas (traditionally air), (b) dielectric elastomer actuators (DEAs) when the dielectric is an elastic solid, (c) soft electrohydraulic actuators (SEHAs) when the dielectric is an uncharged liquid, and (d) electro-kinetic pumps (EKPs) when the dielectric is an electrically charged liquid. To better focus on HVEAs' unique design space, our survey omits high-voltage non-electrostatic actuators, such as piezoelectrics and liquid crystal elastomers, and low-voltage electrostatic actuators, such as conductive polymer networks \cite{Wang_Sotzing_Lee_Chortos_2023}, ionic polymer-metal composites \cite{Feng_Hou_2018}, and stimuli-responsive hydrogels \cite{Paschew_Richter_2010}.
We conducted a systematic literature review following PRISMA reporting guidelines \cite{Page_McKenzie_Bossuyt_Boutron_Hoffmann_Mulrow_Shamseer_Tetzlaff_Akl_Brennan_2021} to understand the research landscape for integrating HVEAs into haptic user interfaces. We filtered 1232 papers from seven databases (Web of Science, IEEE Xplore, ACM Digital Library, Wiley Online Library, Elsevier ScienceDirect, Nature Portfolio, and Science) that included keywords in their metadata related to HVEAs and haptics. Specific search terms and PRISMA flowcharts are included in the Supplementary Materials, Sec. S1. We then manually screened for relevant studies that tested their device on people, either informally on just the authors or through a formal user study. We also annotated each study's haptic interactions and safety considerations. Using this process, we found 173 relevant studies using ESAs, 127 studies using DEAs, 22 studies using SEHAs, and 10 studies using EKPs. 

In this section, drawing from our survey analysis, we first compare HVEAs quantitatively and qualitatively against traditional haptic actuators to understand their design space. We then describe each HVEA's high-level actuation principles, typical configurations, and typical mechanisms for haptic applications. Additional engineering details, including equations for force output, displacement, and capacitance, are included in the Supplementary Materials, Sec. S2.

\subsection{Comparing HVEAs Against Traditional Haptic Actuators} \label{subsec:pros_cons_hveas}

\begin{figure*}
    \includegraphics[width=0.49\linewidth]{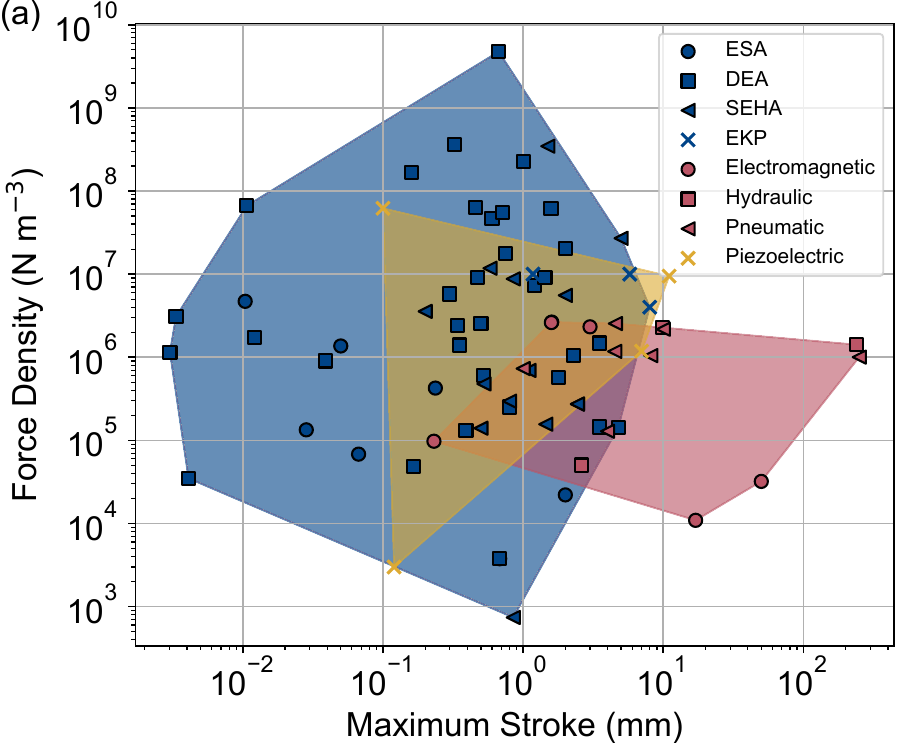}
    \includegraphics[width=0.49\linewidth]{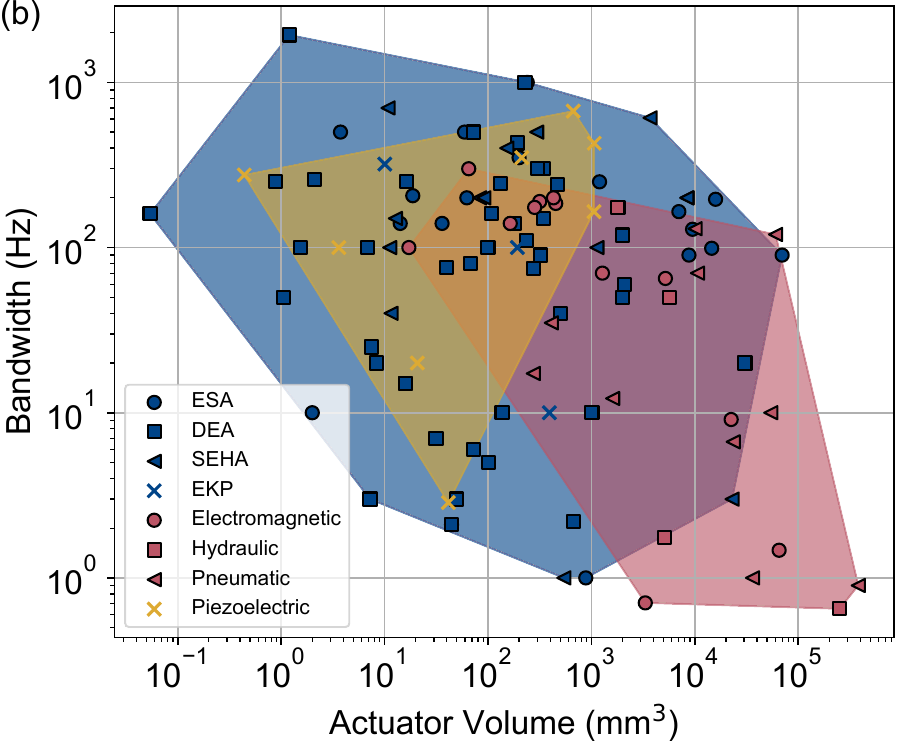}
    \caption{Comparison of performance metrics for actuators used to produce normal skin deformation in prior haptics literature. Actuators were grouped into three broad categories: HVEAs (blue), traditional actuators (pink), and piezoelectrics (yellow). (a) The x-axis shows each actuator's maximum stroke\revision{, defined as its maximum normal displacement without any applied load}, and the y-axis shows the ratio between each actuator's maximum force output and its volume \revision{when producing that force (including fluid volume for fluidic actuators such as SEHAs, EKPs, hydraulics, and pneumatics)}. (b) The x-axis shows each actuator's volume\revision{, defined the same as in (a)}, and the y-axis shows the -3 dB bandwidth of its stroke\revision{, defined as the frequency after which its normal displacement drops below half its maximum stroke}. The raw data and additional data visualizations are provided in the Supplementary Materials, Sec. S3.A.}
    % \caption{Comparison of performance metrics for actuators used to produce normal skin deformation in prior haptics literature. Actuators were grouped into three broad categories: HVEAs (blue), traditional actuators (pink), and piezoelectrics (yellow). (a) The x-axis shows each actuator's maximum stroke, and the y-axis shows the ratio between each actuator's maximum force output and its volume. (b) The x-axis shows each actuator's volume, and the y-axis shows the -3 dB bandwidth of its stroke. For fluidic actuators (SEHAs, EKPs, hydraulic, pneumatic), we include the fluid volume \revision{required to reach maximum displacement} into the plotted actuator volume. The raw data and additional data visualizations are provided in the Supplementary Materials, Sec. S3.B.}
    \label{fig:lit_review_cutaneous_feedback}
\end{figure*}
% Stroke is defined as the maximum normal deformation reported by each actuation system, the maximum force is defined as the maximum active force produced (rather than the maximum blocked force, which some haptics papers also report), the bandwidth as the frequency after which its normal displacement drops below half its maximum stroke, the actuator volume is defined as the volume required to reach maximum force production (including fluid volume, for fluidic actuators), and the force density is calculated by dividing the maximum force by the actuator volume.

% Although prior literature reviews have compared HVEA performance against traditional actuators within the context of robotics \cite{Ankit_Ho_Nirmal_Kulkarni_Accoto_Mathews_2022, Perera_Liyanapathirana_Gargiulo_Gunawardana_2024, Rothemund_Kellaris_Mitchell_Acome_Keplinger_2021a, Vermes_Czigany_2020}, prior haptics surveys have primarily focused on qualitative comparisons \cite{ Yin_Hinchet_Shea_Majidi_2021, Bai_Li_Shepherd_2021}. 

Fig. \ref{fig:lit_review_cutaneous_feedback} compares performance metrics between HVEAs and traditional actuators used to produce normal skin deformation in prior haptics literature. \revision{Actuator classes were drawn from prior literature reviews on haptic actuators \cite{Ankit_Ho_Nirmal_Kulkarni_Accoto_Mathews_2022, Yin_Hinchet_Shea_Majidi_2021, culbertson2018haptics, Bai_Li_Shepherd_2021}.}

\revision{Fig. \ref{fig:lit_review_cutaneous_feedback}(a) plots each actuator's maximum stroke against the ratio between its maximum force output and its volume when producing that force. For fluidic actuators (SEHAs, EKPs, hydraulic, pneumatic), we include the fluid volume into the calculated actuator volume.} As shown, HVEAs deform less than electromagnetic, pneumatic, or hydraulic actuators, but in return HVEAs offer exceptional output force \revision{density}. This high force density is because electrostatic forces scale with surface area, and thus the ratio between electrostatic force ($F \propto l^2$, where $l$ is a symbolic length dimension) and volume ($V \propto l^3$) improves at small sizes. By contrast, the force output of many traditional actuator technologies scales with the volume of their component magnets, pumps, or fluid chambers, explaining why they offer better forces and displacements than HVEAs at larger sizes. 

% As shown in Fig. \ref{fig:lit_review_cutaneous_feedback}(a), HVEAs deform less than electromagnetic, pneumatic, or hydraulic actuators, but in return HVEAs offer exceptional output force \revision{density}. This high force density is because electrostatic forces scale with surface area, and thus the ratio between electrostatic force ($F \propto l^2$, where $l$ is a symbolic length dimension) and volume ($V \propto l^3$) improves at small sizes. By contrast, the force output of many traditional actuator technologies scales with the volume of their component magnets, pumps, or fluid chambers, explaining why they offer better forces and displacements than HVEAs at larger sizes. 
% Piezoelectric actuators offer a compromise between these size regimes \cite{Xie_Liu_Yang_Yang_Liu_Xu_Zhang_Zhai_2017}, and we discuss in the Supplementary Materials, Sec. S1.B how many flexible piezoelectric actuators can also function as DEAs given a large enough drive voltage.

Fig. \ref{fig:lit_review_cutaneous_feedback}(b) \revision{plots each actuator's volume against its dynamic bandwidth, highlighting a roughly inverse relationship.} Because of their high force density, HVEAs can be miniaturized to improve their bandwidth while still providing enough force output for haptics applications. Almost every HVEA in our survey had response times on the order of milliseconds, offering a strong promise for dynamic haptic interaction even while approaching the spatial resolution limit of human perception. 

Beyond quantitative metrics, HVEAs can be readily fabricated using monolithic manufacturing techniques such as printed circuit board manufacturing, silicone molding, and laser cutting, which allows researchers to scalably manufacture high-resolution actuator arrays. In applications such as refreshable braille displays \cite{Runyan_Blazie_2011} and shape displays \cite{Rauf_Bernardo_Follmer_2023, Zhang_Gonzalez_Guo_Follmer_2019}, this scalability allows HVEAs to achieve order-of-magnitude improvements in cost and manufacturing complexity compared to traditional actuators. HVEAs' versatile manufacturing process also means that every HVEA can be implemented in flexible form factors, enabling more ergonomic form factors for wearables and more seamless integration into world environments. We discuss these fabrication workflows further in Sec. \ref{subsec:fabricating_hveas}.

HVEAs also come with several qualitative disadvantages compared to traditional haptic actuators. HVEAs' high voltage requirements demand special safety and circuit design considerations, and commercial high voltage power supplies are often bulky and expensive. Moreover, most HVEAs are not sold off-the-shelf for easy prototyping, and their manufacturing processes, while scalable and theoretically efficient, can sometimes involve specialized materials such as carbon black powder, liquid dielectrics, and porous glass fillers. HVEA performance also tends to degrade in non-ideal conditions, such as in high humidity or dirty environments, although this degradation has not yet been rigorously failure tested to the extent that traditional actuators have. We present recommended design workflows, safety guidelines, and circuit components that address some of these limitations in Sec. \ref{sec:design_workflows}.
% or after daily life events, such as dropping, creasing, or washing the HVEA. . Although the extent of this deterioration can vary drastically, HVEAs have not 

% , such as the piezoelectric actuators used in conventional Braille displays
% \textcolor{red}{Table 1 in this paper might be interesting to be referred to here: https://advanced-onlinelibrary-wiley-com.focus.lib.kth.se/doi/epdf/10.1002/adfm.202007428}

The next four subsections will discuss design trade-offs further for each individual HVEA type, describing each of their high-level actuation principles and haptic mechanisms.

\begin{figure*}
    \centering
    \includegraphics[width=\linewidth]{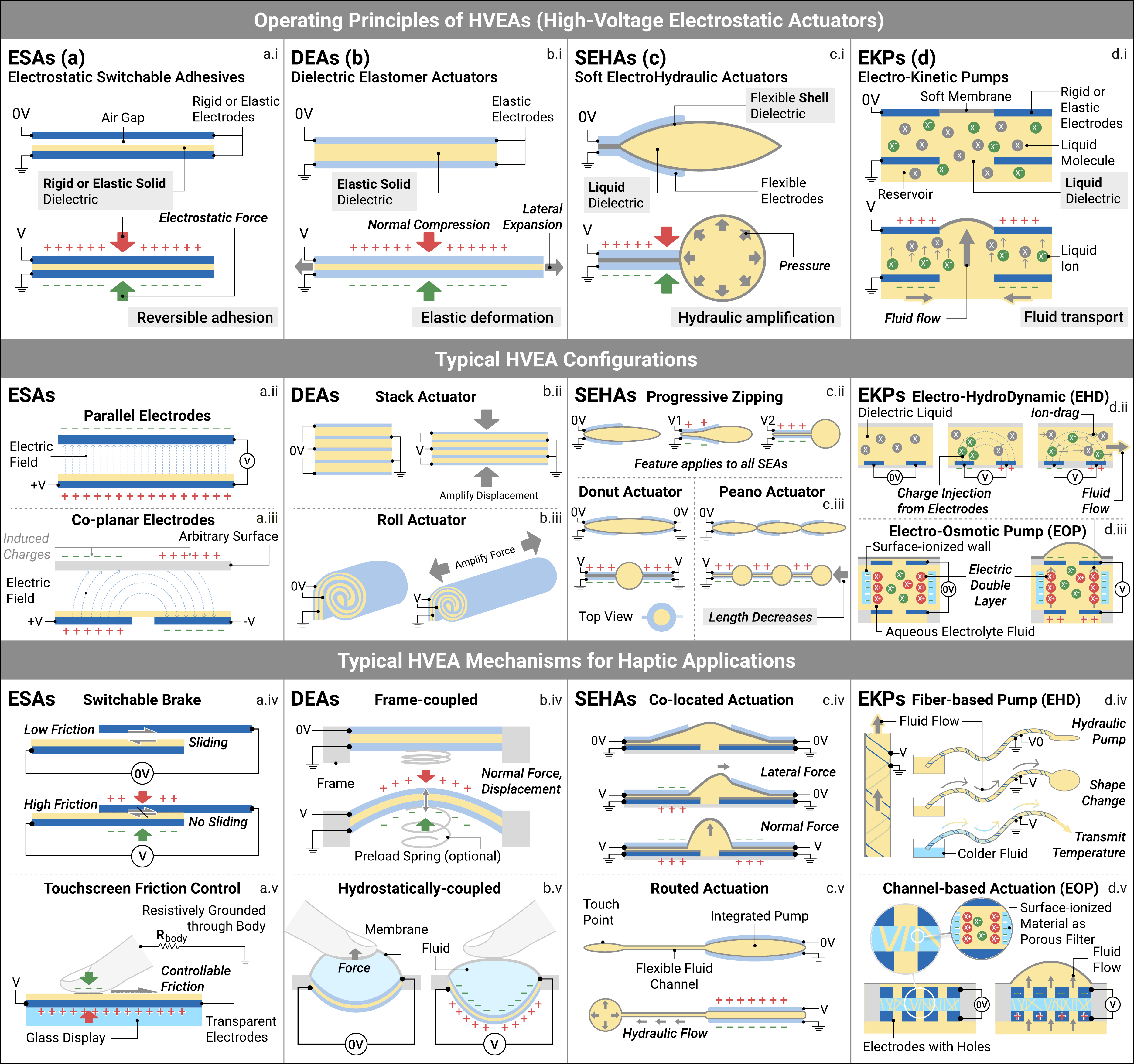}
    \caption{Schematic diagrams of each HVEA type's (i) operating principles, (ii-iii) typical electrode layouts and configurations, and (iv-v) typical mechanisms for haptics applications. Each HVEA's geometry and dielectric material significantly affect its haptic output.}
    % \textcolor{red}{Open to any comments.} \href{https://www.figma.com/design/6thHVgA1fjHfQ7HdOPdw7Z/HVEAs-Paper-Figures?node-id=1-3\&t=JAOOJ4cHSuiDDX6u-1}{Link to Figma diagram here}
    \label{fig:hvea_principles_mechanisms}
\end{figure*}

\vspace{2.5mm}
\subsection{Electrostatic Switchable Adhesives (ESAs)}  \label{subsec:esa_definition}

\begin{figure}
    \includegraphics[width=\linewidth]{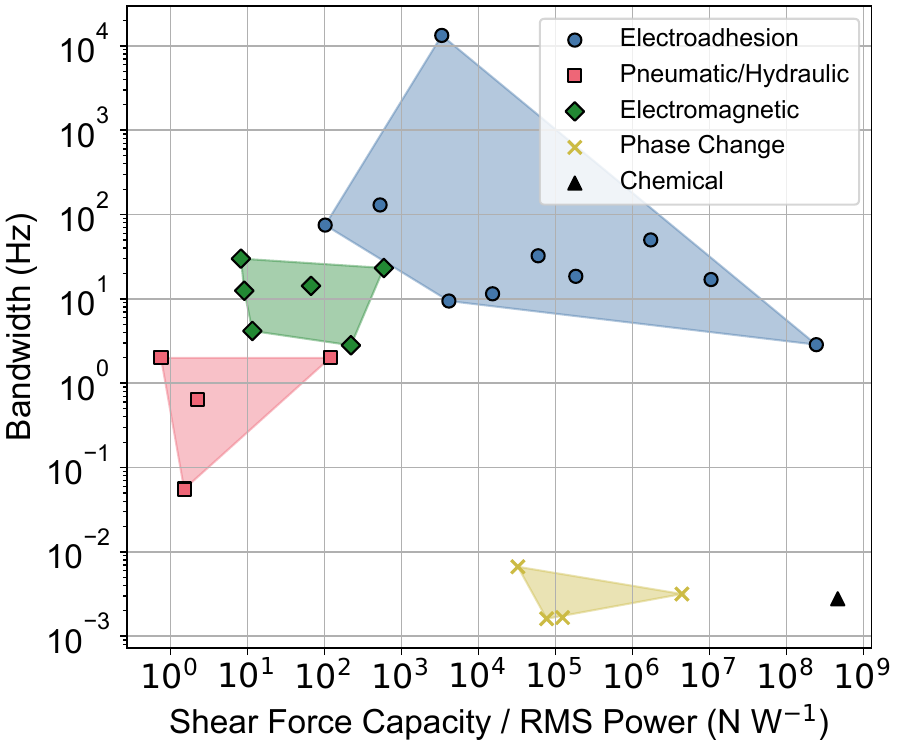}
    \caption{Comparison of performance metrics for centimeter-scale switchable adhesives adhering to bulk material substrates in the literature\revision{, with categories drawn from prior review papers \cite{Croll_Hosseini_Bartlett_2019, Liu_Yan_2022}}. The x-axis shows the ratio between shear force capacity and power consumption, and the y-axis shows the bandwidth, which we define as the drive frequency after which the shear force capacity drops to half its maximum value. \revision{More details about the data collection and} raw data is provided in the Supplementary Materials, Sec. S3.B.}
    \label{fig:lit_review_switchable_adhesives_bandwidth}
\end{figure}
% For papers annotated with an asterisk (*), bandwidth was estimated as the inverse of the engagement time plus the release time. 
% \cite{Zhang_Gonzalez_Guo_Follmer_2019, Shultz_Peshkin_Colgate_2018, Diller_Collins_Majidi_2018, Hinchet_Shea_2020, Li_Xiong_Ma_Yang_Ma_Tao_2023, Najmaei_Asadian_Kermani_Patel_2015, Li_Furusho_Morimoto_Tokuda_Hashimoto_2007, Gavin_Hoagg_Dobossy_2001, Baser_Demiray_Bas_Kilic_Erol_2017, Alkan_Gurocak_Gonenc_2013, Vogel_Steen_2010, Kremer_Nohooji_Sanchez-Lopez_Voos_2023, Kim_Yoon_Kim_Yun_2024, Jiang_Ma_Chen_Liu_Li_Paik_2021, Michal_Spencer_Rowan_2016, Wu_Ji_Zhao_Han_Müllen_Pan_Yin_2019, Ito_Akiyama_Sekizawa_Mori_Yoshida_Kihara_2018, Gou_Hou_Saed_2024, Ahn_Jang_Selvapalam_Yun_Kim_2013}

% Electrostatic adhesion is the electrostatic attraction between two contacting materials separated by a dielectric and held at a potential difference. Its scalable manufacturing process, high shear force capacity against a wide range of substrates, relatively fast actuation speeds, low power consumption, and low profile have made electrostatic switchable adhesives useful for space- and power-constrained applications. 

Electrostatic switchable adhesion, often shortened to electroadhesion, is the electrostatic attraction between two contacting or near-proximity surfaces held at a potential difference. As shown in Fig. \ref{fig:hvea_principles_mechanisms}(a), it is implemented in two primary ways. 

% The bipolar, parallel electrode configuration mimics a traditional electrical capacitor, using parallel electrodes and requiring a power supply that outputs two separate voltages, typically a high voltage $V$ and ground. Applying this potential difference between the parallel electrodes causes each electrode to accumulate opposite charges that then attract the two surfaces together.
The parallel electrode configuration mimics a traditional electrical capacitor. Applying a potential difference between the electrodes causes them to accumulate opposite charges that then attract the two surfaces together. Although both electrodes are usually directly driven by the power supply, if one surface (most commonly a human fingertip in haptics applications) is semiconducting and resistively connected to earth ground, the power supply can just drive the other surface with a voltage or current also referenced to earth ground. This latter case forms a resistive divider between the dielectric's resistance, the fingertip-to-dielectric contact resistance, and the body's resistance to earth ground. The contact resistance's potential drop generates an electrostatic force, called the Johnsen-Rahbek effect \cite{Johnsen_Rahbek_1923}, that attracts the fingertip to the dielectric.

In the co-planar electrode configuration, potential differences between the interdigitated electrodes patterned onto the dielectric create fringing fields that induce an equal but opposite charge distribution in the opposing surface. Similar to the Johnsen-Rahbek configuration, co-planar electrodes allow all the electronics to be concentrated into a single surface, reducing bulk and enabling the clutch to adhere to generic surfaces in the field and to small, delicate objects. However, because the second surface is not connected to the main circuit, the potential difference between surfaces is only half the power supply's, decreasing the attractive force by 75\% compared to the parallel electrode configuration \cite{Rauf_Follmer_2025}.
% , and the other surface can be a dielectric or conductor

% The third unipolar, parallel electrode configuration uses a single electrode and a current source instead of a voltage source. It requires the second surface, typically a human fingertip in haptics contexts, to start off in contact with the dielectric-coated electrode. When a current is sent into the electrode, a resistive divider forms between the dielectric's resistance, the fingertip-to-dielectric contact resistance, and the fingertip's resistance to earth ground. The contact resistance's potential drop generates an electrostatic force, called the Johnsen-Rahbek effect after its discoverers in 1923 \cite{Johnsen_Rahbek_1923}, that attracts the fingertip to the dielectric. Although the first two ESA configurations work better for bulk conductive or insulating substrates, this configuration generates exceptional attractive forces to human skin because bound ions in biological tissue amplify the electric field in what is known as the ``electric double layer'' effect \cite{Kuang_Nelson_1998}.
% As a technical note, the electrode's dielectric coating is important for producing Johnsen-Rahbek electroadhesion. Without the dielectric, the stimuli would be better characterized by electrotactile stimulation \cite{Kourtesis_Argelaguet_Vizcay_Marchal_Pacchierotti_2022}.
% AliAbbasi_Martinsen_Pettersen_Colgate_Basdogan_2024, 

ESAs are unique among HVEAs for their ability to adhere to nearby objects with only microscopic deformation. The resulting friction from this normal force can also be used to counteract applied shear loads on either surface, and it can be coupled with an elastic actuator or spring for variable stiffness control or shape change. However, these mechanisms are often space inefficient, and as shown in Fig. \ref{fig:lit_review_cutaneous_feedback}(a) ESA cutaneous feedback mechanisms generally have the smallest output deformation and force density among HVEAs. ESA electrodes can be patterned onto a wide variety of substrates, from touchscreens \cite{Bau_Poupyrev_Israr_Harrison_2010} to flexible printed circuit boards \cite{Rauf_Bernardo_Follmer_2023} to textiles \cite{Vechev_Hinchet_Coros_Thomaszewski_Hilliges_2022}. As shown in Fig. \ref{fig:lit_review_switchable_adhesives_bandwidth}, ESAs achieve response times and shear force capacities several orders of magnitude higher than other switchable adhesion mechanisms, enabling dynamic, energy-efficient kinesthetic haptic feedback.
% Most ESAs for haptics applications also work well using voltages under 1 kV, simplifying circuit design as discussed further in Sec. \ref{sec:design_workflows}. 
% Its scalable manufacturing process, high shear force capacity against a wide range of substrates, fast (millisecond-scale) actuation speeds, low power consumption, and low profile (often $<$ 100 \textmu m) make ESAs useful for space- and power-constrained applications. 
% Literature reviews on switchable adhesion \cite{Ko_Javey_2017}

\subsection{Dielectric Elastomer Actuators (DEAs)}  \label{subsec:dea_definition}
Dielectric elastomer actuators (DEAs) are a class of electroactive polymer that change their shape in response to an applied electric field. Originally discovered in 1880 by Röentgen after spraying charges onto rubber \cite{Rontgen_1880}, DEAs are commonly implemented today by sandwiching a soft dielectric between compliant electrodes. As illustrated in Fig. \ref{fig:hvea_principles_mechanisms}(b.i), when a voltage is applied between the electrodes, the dielectric compresses, decreasing in thickness and increasing in area. 

DEAs can output large strains (10-50\%) and moderate stresses ($\sim$100 kPa) within a compact form factor, and their simple layer stack allows them to be scalably fabricated in high-spatial-resolution arrays. As shown in Fig. \ref{fig:hvea_principles_mechanisms}(b.iv), many designs use a pre-stretched DEA film fixed to a rigid boundary frame. When a voltage is applied, the membrane's lateral expansion is constrained by the frame, causing it to buckle out of plane and generate haptic feedback. Preload springs can also be added to amplify this displacement \cite{youn2025skin}. Shear deformation can also be rendered by fixing the DEA to only one side of the boundary frame \cite{Han_Bae_Gregoriou_Ploch_Goldman_Glover_Daniel_Cutkosky_2018}. As shown in Fig. \ref{fig:hvea_principles_mechanisms}(b.v), these DEAs can be used as a deformable membrane on a fluid-filled pouch \cite{frediani2014wearable}. The hydrostatically coupled force can be felt by a user interacting with the pouch's non-DEA surfaces, allowing the user to avoid direct contact with the high-voltage electrodes. 
% Fig. \ref{fig:hvea_principles_mechanisms} shows several typical haptic mechanisms. 
% , allowing an end user to directly feel the DEA's lateral expansion
% When a voltage is applied, the DEA membrane expands and buckles outward, drawing fluid away from the rest of the pouch. 

Among HVEAs, DEAs stand out for their exceptional force density and bandwidth, but as shown in Fig. \ref{fig:lit_review_cutaneous_feedback}(a, b) their performance drops drastically when scaled to larger sizes. To address this, DEAs can be stacked in series or parallel to enhance their output. Stacking DEAs in series (Fig. \ref{fig:hvea_principles_mechanisms}(b.ii)) amplifies their normal deformation proportional to the number of layers. Alternatively, a large DEA sheet can be rolled into a tight cylinder that expands axially and shrinks in diameter once a voltage is applied (Fig. \ref{fig:hvea_principles_mechanisms}(b.iii)). This effectively stacks each DEA layer in parallel, and the cylinder's force output scales proportional to the number of layers. Care must be taken when stacking or rolling DEAs, however, because the dielectric breakdown of a single DEA can propagate through and destroy the entire stack. We discuss how to add a safety factor to address this risk in Sec. \ref{subsec:high_voltage_safety}.

\subsection{Soft ElectroHydraulic Actuators (SEHAs)} \label{subsec:sea_definition}
% I want to write a version whose flow follows to our figure structure better and shorter. If this makes sense, we can apply it to other technology too. 

Soft ElectroHydraulic Actuators (SEHAs), originally introduced as Hydraulically Amplified Self-healing ELectrostatic (HASEL) actuators \cite{acome2018hydraulically}, consist of a flexible shell partially covered by electrodes and filled with a liquid dielectric (Fig. \ref{fig:hvea_principles_mechanisms}(c.i)). When voltage is applied, the resulting electrostatic force squeezes the shell shut, displacing the liquid from between the electrodes into the surrounding volume. Because liquids are nearly incompressible, this hydraulic flow can be efficiently directed from a large pouch to a small end effector, amplifying force output and enabling larger shape change.

% As illustrated in Fig. \ref{fig:hvea_principles_mechanisms}(c.ii), SEAs progressively zip from the shell's edge towards its center as the applied voltage increases. As illustrated in Fig. \ref{fig:hvea_principles_mechanisms}(c.iii), SEAs are commonly designed to either expand in thickness or contract in length. 
As illustrated in Fig. \ref{fig:hvea_principles_mechanisms}(c.ii, c.iii), SEHAs progressively zip from the shell's edge towards its center as the applied voltage increases, expanding in thickness and contracting in length. Expanding SEHAs, such as donut actuators, can achieve large strains ($>$100\%), while contracting SEHAs, such as Peano-HASEL actuators, can output large forces ($>$100 kPa) \cite{Kellaris_Gopaluni_Venkata_Smith_Mitchell_Keplinger_2018}.
% SEAs operate through progressive zipping (Fig.~\ref{fig:hvea_principles_mechanisms}(c.ii)), which begins at the shell edge where the electrodes are closest and the electric field is strongest. SEAs are commonly designed to either expand in thickness or contract in length. Examples of these common layouts are donut and Peano actuators (Fig.~\ref{fig:hvea_principles_mechanisms}(c.iii)), respectively. Expanding SEAs can achieve large actuation strain ($>$100\%) while contracting SEAs can output large actuation stress ($>$100 kPa) \cite{Kellaris_Gopaluni_Venkata_Smith_Mitchell_Keplinger_2018}.

% The two common layouts are donut and Peano actuators (Fig.~\ref{fig:hvea_principles_mechanisms}(c.iii)). Donut actuators use ring-shaped electrodes to create a large central bump, while Peano actuators connect multiple pouches in series to amplify total length change \cite{Kellaris_Gopaluni_Venkata_Smith_Mitchell_Keplinger_2018}. 

Most haptic SEHAs incorporate co-located actuation, where liquid is redistributed within a single pouch \cite{leroy2020multimode, grasso2023fully}. As shown in Fig.~\ref{fig:hvea_principles_mechanisms}(c.iv), a variant of this design, called Hydraulically Amplified taXELs (HAXELs), can also leverage multiple electrodes per pouch to control whether the force output is primarily normal or shear \cite{leroy2020multimode}. The incompressibility of liquids also enables routed actuation (Fig.~\ref{fig:hvea_principles_mechanisms}(c.v)), in which the actuation source is separated from the output inflatable pouch either vertically \cite{han2020haptic, chen2023skin, sirbu2019electrostatic} or via integrated fluid channels \cite{purnendu2023fingertip, shao2025wearable}, allowing a more compact footprint at the contact point.
% employ independently controllable electrodes. Engaging the electrodes entirely pushes all the liquid into the center, creating a strong normal force, while partially activating the electrodes pushes the liquid to the opposite side, producing a combined normal and shear force

Unlike the solid dielectric in DEAs, the liquid dielectric in SEHAs can reflow after dielectric breakdown to restore insulation, providing self-healing and improving device lifetime and reliability. This property allows stacked SEHAs to achieve large deformations with less risk of catastrophic dielectric failure if a single SEHA in the series breaks down \cite{acome2018hydraulically, Johnson_Naris_Sundaram_Volchko_Ly_Mitchell_Acome_Kellaris_Keplinger_Correll_2023}.

\subsection{Electro-Kinetic Pumps (EKPs)} \label{subsec:ekp_definition}
Electro-Kinetic Pumps (EKPs) arrange electrodes within a dielectric liquid. They generate fluid flow by injecting \revision{charge} into the dielectric and then attracting the charged liquid towards the powered electrodes. Unlike SEHA electrodes, which are insulated from the dielectric through the elastomeric pouch, EKP electrodes make direct contact with the liquid. Although charges can be injected in different ways \cite{Iverson_Garimella_2008}, haptic systems have primarily explored two methods: Electro-HydroDynamic \revision{Pumps} (EHDPs) and Electro-Osmotic Pumps (EOPs).

% Electro-Kinetic Pumps (EKPs) work by arranging high voltage and ground electrodes within a dielectric liquid. Unlike SEHAs, where electrodes are insulated from the dielectric through the elastomeric pouch, EKP electrodes make direct contact with the liquid. Fluid flow is generated by injecting \revision{charge} into the dielectric and then attracting the charged liquid towards the powered electrodes. Although charges can be injected in different ways \cite{Iverson_Garimella_2008}, haptic systems have primarily explored two methods: Electro-HydroDynamic \revision{Pumps} (EHDPs) and Electro-Osmotic Pumps (EOPs).
% local charge concentrations

EHDPs inject electrons directly from the negatively charged electrode by applying a large voltage between closely spaced electrodes (Fig.~\ref{fig:hvea_principles_mechanisms}(d.ii)). These ions migrate toward the positive electrode, dragging the dielectric fluid along. EHDPs tend to generate low static pressures (usually 5-10 kPa) and high flow rates ($>$ 50 mL/min) \cite{Smith_Cacucciolo_Shea_2023, Morita_Kuwajima_Minaminosono_Maeda_Kakehi_2022}, making them useful for fluid transport in fiber-based pumps (Fig.~\ref{fig:hvea_principles_mechanisms}(d.iv)).  

EOPs take advantage of the fact that free charges inherently form at solid-liquid interfaces by including a porous filter soaked in dielectric liquid as a spacer between the electrodes (Fig.~\ref{fig:hvea_principles_mechanisms}(d.iii \& d.v)). This porous filter increases the number of solid-liquid interfaces and thus the number of free charges, which allows EOPs to operate at much lower voltages than EHDPs. From their structure, EOPs tend to generate higher pressures ($>$ 50 kPa) and lower flow rates (around 5-10 mL/min) than EHDPs \cite{Shultz_Harrison_2023}, making them well-suited for haptic applications that utilize shape change and normal force.

Functionally, EKPs combine the high spatial resolutions of DEAs with the self-healing liquid dielectric advantages of SEHAs. Their form factor is less versatile than that of DEAs and SEHAs due to the need for precisely spaced electrodes (in the case of EHDPs) or porous filters (in the case of EOPs), but we envision future work will explore novel mechanisms that take advantage of EKPs' unique properties.

% Their rigid form factor lacks the versatility of DEA geometries and limits the potential for hydraulic amplification like with SEAs, however.

%% file: 3_integrating_electrostatic_actuators_into_haptic_systems.tex
\section{Integrating Electrostatic Actuators Into Haptic Systems}
\label{sec:integrating_hveas_into_haptic_systems}

\revise{Human haptic perception is derived from sensory organs distributed across the body and is commonly divided into two modalities: cutaneous and kinesthetic \cite{lederman2009haptic}. Cutaneous haptics is perceived through mechanoreceptors, thermoreceptors, and nociceptors (pain receptors) embedded in the skin to convey localized information such as pressure, texture, and vibration. Kinesthesia is defined as the body's sense of movement and position, and kinesthetic feedback is produced by applying forces and torques that can be sensed in a user's muscles, tendons, and joints. Inspired by Culbertson et al. \cite{culbertson2018haptics}, we also define haptic surfaces, which allow users to explore haptic actuators distributed across a large surface for simultaneous cutaneous and kinesthetic feedback interactions.}

% Accordingly, we structure this section by categorizing
To understand the design space of prior work integrating HVEAs into haptic systems, we categorize each paper in our survey by whether its intended purpose was to produce localized cutaneous feedback, kinesthetic feedback, or a haptic surface. We also denoted whether the haptic device was active (contributing energy or outputting positive work on the user) or passive (dissipating energy provided by the user or system), in line with previous wearable haptics taxonomies \cite{Pacchierotti_Sinclair_Solazzi_Frisoli_Hayward_Prattichizzo_2017, Dahiya_Metta_Valle_Sandini_2010}.

\subsection{Localized Cutaneous Feedback} \label{sub:localized_cutaneous_feedback}

We categorize localized cutaneous feedback actuators based on their ability to stimulate skin mechanoreceptors responsive to normal skin deformation and lateral skin stretch. We include vibrotactile feedback as an optional addition to these two categories if HVEAs trigger this skin deformation at high frequency. Several studies also explored cutaneous feedback stimuli other than skin deformation, which we discuss in Sec. \ref{subsubsec:miscellaneous_cutaneous_feedback}. Within each group, we further organize the devices by their form factors, technologies, and applications. Fig. \ref{fig:lit_review_images_cutaneous} highlights several examples of previous HVEA haptic integrations targeted at localized cutaneous feedback. \revision{We include more details about human haptic perceptual thresholds of localized cutaneous feedback in the Supplementary Materials, Sec. S3.}
% Cutaneous feedback beyond skin deformation is grouped under miscellaneous interactions.
% We include vibrotactile feedback as an optional feature within these two categories because the majority of HVEAs' localized vibration effects are produced through high-frequency operation of these deformation mechanisms. 
% Many HVEAs also inherently possess high bandwidths in the hundreds to thousands of Hertz, allowing them to target multiple different mechanoreceptors at different frequency ranges. 
% Fig. \ref{fig:lit_review_images_cutaneous} highlights several examples of previous HVEA haptic integrations, divided by their form factors.

\subsubsection{\revision{Normal Skin Deformation with Optional Vibrotactile Feedback (Wearables)}}
\label{subsubsec:normal_skin_deformation}
Actuators that provide normal deformation and force to the skin enable users to perceive contact, shape, and dynamic pressure. These cues can enable applications such as simulating physical object encounters in VR or rendering salient social and affective touch. Our survey identified actuators that produce these cues by deforming the skin through either discrete on-off or continuous pressure modulation, sometimes accompanied by low- or high-frequency vibration. Among HVEAS, normal skin deformation is most commonly achieved through out-of-plane (OOP) actuation, with an alternative approach using bending DEAs that operate similarly to rotary motors \cite{cameron2011bending}. We found OOP actuators in 26 ESA papers, 83 DEA papers, 15 SEHA papers, 5 EKP papers, and 4 papers that combine DEAs and ESAs. 
HVEAs share an advantage of being thin and compact, making them well-suited for high-resolution OOP actuation. Different technologies, however, show distinct trade-offs: DEAs and ESAs generally benefit from simpler layer stacks that support finer spatial resolutions, while SEHAs and EKPs tend to deliver larger displacements and forces. For instance, Pyo et al. used DEAs to create the highest spatial resolution actuator array in our literature review, a 10x10 array spaced at 400 pixels/cm$^2$, but each actuator only had a stroke of 10 \textmu m and a force output of 3.6 mN \cite{Pyo_Ryu_Kyung_Yun_Kwon_2018}. The highest-resolution ESA array in our survey had 44.4 pixels/cm$^2$, but each dot could only output 60 \textmu m stroke and 2.5 mN output force \cite{Takagi_Sasaki_Shikida_Sato_2007}. The highest-resolution SEHA array had 16 pixels/cm$^2$ \cite{purnendu2023fingertip}, using routed hydraulic channels to separate the higher-resolution output array from the bulkier electrodes and fluid reservoirs. Its actuators could displace up to 1.66 mm and output forces up to 60 mN. The highest-resolution EKP dot array had 20 pixels/cm$^2$, and each pixel had a 120 \textmu m stroke and 100 mN force output \cite{Shen_Rae-Grant_Mullenbach_Harrison_Shultz_2023}. \revise{Rolled DEAs can also be used for high-force applications, although they tend to be less compact than similar SEHAs or EKPs; for example, Levard et al. \cite{Levard_Diglio_Lu_Gorny_Rahn_Zhang_2011} built an array with 16 pixels/cm$^2$ in which each DEA could output displacements up to 1.2 mm and block forces over 1 N.}

\begin{figure}[t]
    \centering
    \includegraphics[width=\linewidth]{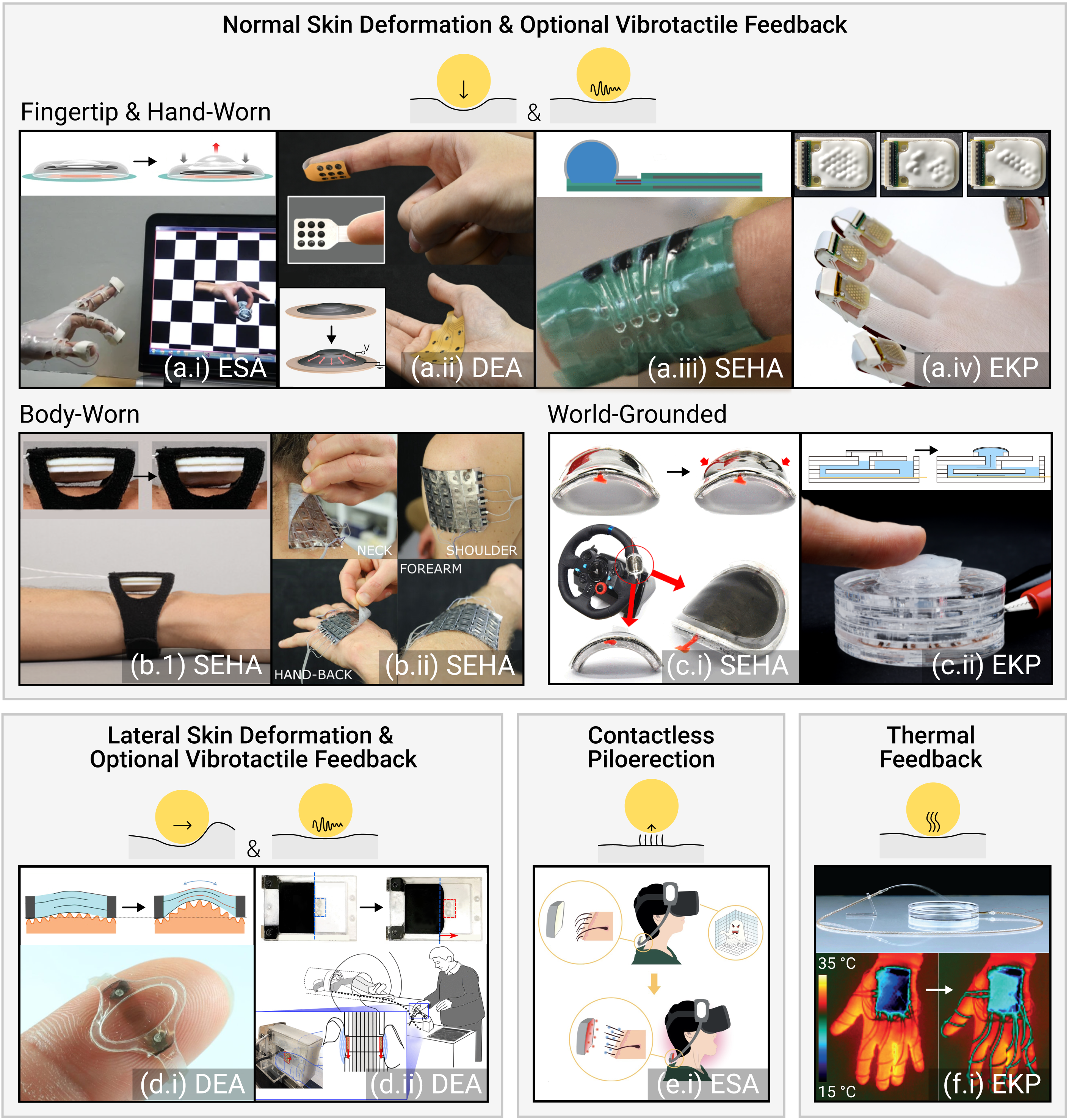}
    \caption{Examples of HVEA haptic applications targeting localized cutaneous feedback. The images were drawn from (a.i) \cite{Song_Kim_Jin_Kim_Lee_Kim_Park_Cha_2019} (\copyright 2019 The Authors), (a.ii) \cite{Youn_Jang_Hwang_Pei_Yun_Kyung_2025} (\copyright 2025 The Authors), (a.iii) \cite{purnendu2023fingertip} (\copyright 2023 IEEE), (a.iv) \cite{Shen_Rae-Grant_Mullenbach_Harrison_Shultz_2023} (\copyright 2023 The Authors), (b.i) \cite{sanchez2024cutaneous} (\copyright 2024 The Authors. Advanced Science published by Wiley‐VCH GmbH), (b.ii) \cite{Leroy_Shea_2023} (\copyright 2023 The Authors. Advanced Materials Technologies published by Wiley‐VCH GmbH), (c.i) \cite{Kim_Nam_Kim_Kyung_2021} (\copyright 2021 IEEE), (c.ii) \cite{fujii2021layerpump} (\copyright 2021 The Authors), (d.i) \cite{Ji_Liu_Cacucciolo_Civet_ElHaitami_Cantin_Perriard_Shea_2021} (\copyright 2020 Wiley‐VCH GmbH), (d.ii) \cite{Han_Bae_Gregoriou_Ploch_Goldman_Glover_Daniel_Cutkosky_2018} (\copyright 2018 IEEE), (e.i) \cite{Iriarte_Ezcurdia_Elizondo_Irisarri_Hemmerling_Ortiz_Marzo_2024} (\copyright 2024 IEEE), (f.i) \cite{Smith_Cacucciolo_Shea_2023} (\copyright 2023 The American Association for the Advancement of Science).  Icons inspired by \cite{frisoli2024wearable}.}
    \label{fig:lit_review_images_cutaneous}
\end{figure}

% \paragraph{Wearable Devices}

Wearable devices represent an important class of haptic interfaces, as they provide effective feedback on the skin while allowing users to move freely in physical space. However, designing wearable haptic devices that are both expressive and unobtrusive remains challenging \cite{Ankit_Ho_Nirmal_Kulkarni_Accoto_Mathews_2022, Yin_Hinchet_Shea_Majidi_2021}. HVEAs can help address long-standing challenges in wearable haptics, as they can be made soft, thin, and flexible to conform to the human body. These have been made using all four HVEA technologies, and we broadly categorize prior work into single actuators and pixel arrays. 
% , while remaining self-contained and miniaturized. The papers we reviewed demonstrate wearable devices integrating HVEAs for normal skin deformation across all four technologies
% , for various body placements

% \textit{Single-unit Wearable Devices} 
% \textit{Wearable Single Actuators}
\textit{Single-Unit Wearable Actuators} are primarily applied to the fingertip to deliver normal force and vibrotactile feedback in Virtual Reality (VR). As shown in Fig. \ref{fig:lit_review_images_cutaneous}(a.i), Song et al. \cite{song2019pneumatic} developed a soft ESA actuator that operates like a pneumatic version of a donut SEHA, producing gentle on-off cues when integrated in a glove. Mun et al. \cite{mun2018EAP} integrated a flexible frame-coupled stacked DEA actuator into a rubber glove, enabling controllable surface protrusions and high-frequency vibrations to simulate cues such as button clicks, localized elastic response, and bubble bursts in VR. Several studies have also explored mounting rigid-coupled \cite{youn2021wearable} and hydrostatically-coupled DEAs \cite{boys2018soft} near the fingertips using rigid structural parts. These designs structurally support the actuator to press into the skin when activated, improving responsiveness but reducing the device's compactness and compliance. Shao et al. \cite{shao2025wearable} introduced a routed SEHA design that positions the soft pumping pouch on the back of the hand and transmits dielectric liquid through thin, flexible tubes to a 30-\textmu m-thick fingertip chamber. This configuration provides strong, salient feedback, while keeping the fingertip unit unobtrusive. In their VR demonstrations, the actuator could render natural touch sensations such as pressing on soft objects and perceiving the geometry of virtual objects during grasping. 
% providing effective on-off touch, µ
% (typically 15 mm in diameter to fit the finger pad) 

Single-unit wearable actuators have also been designed for the arm. Sanchez‐Tamayo et al. designed CUTE \cite{sanchez2024cutaneous}, a co-located SEHA worn on the wrist like a watch (Fig. \ref{fig:lit_review_images_cutaneous}(b.i)). In their perceptual study, participants could accurately distinguish between different touch patterns and described the sensations as pleasant and expressive, ranging from calming to exciting. High-frequency cues were associated with alarms, while more regular patterns with slow transitions and pauses were described as lifelike. 
% , capable of producing sufficient force and displacement to be salient on the hairy skin

% It can output 2.44 mm displacement, more than 2.3 N force, and a bandwidth from 0-200 Hz. Users reported that it provides pleasant and expressive haptic sensations to hairy skin on the arm, and a follow-up work applied the device for button and slider interactions in virtual reality menus \cite{bartels2024active}.

\textit{Wearable Pixel Arrays} extend single actuators by enabling programmable spatiotemporal control. This allows them to render finer spatial details for realistic simulations of shape contours, surface texture, and complex dynamic patterns, making them especially valuable for VR applications. Our review identified 14 papers introducing wearable HVEA-based pixel arrays, demonstrating a range of form factors and mechanisms.
% offering higher spatial resolution and 

Youn et al. \cite{youn2025skin} developed an ultra-thin (1.1 mm) 3x3 DEA array that could produce patternable tactile feedback on the fingertip, palm, and forearm, as shown in Fig. \ref{fig:lit_review_images_cutaneous}(a.ii). Using a compression spring mechanism, it achieves high force density outperforming many earlier DEA arrays \cite{Koo_Jung_Koo_Nam_Lee_Choi_2008, guo2024haptic}, and each pixel could block forces up to 2.2 N and output displacements up to 310 \textmu m at 280 Hz. The device could render both low-frequency tactile patterns and high-frequency vibration, and user studies demonstrated that participants could reliably discriminate dynamic patterns, frequencies, and textures.
% , indicating the array’s capability to render versatile tactile cues with high quality.
%The device also integrates a photomicrosensor array, enabling bi-directional wireless tactile communication \cite{youn2025skin}, which could greatly enrich XR experiences. 
% could be perceived well across the fingertip, palm, and forearm

HAXEL SEHAs, as described in Sec. \ref{subsec:sea_definition}, allow systems to dynamically switch between normal and shear force output. First introduced in 2020 by Leroy et al. \cite{leroy2020multimode}, HAXELs have since been manufactured with silicone for improved compliance and wearability \cite{grasso2023fully} and in various sizes from 3 mm (for fingertip arrays) to 15 mm (for arrays on the hand, forearm, shoulder, neck, or lower back) \cite{Leroy_Shea_2023}. These studies report that users can clearly distinguish different spatio-temporal cues, including moving rows of raised pixels to signify a ``swipe,'' rotating patterns in clockwise and anti-clockwise directions, and rendering high-frequency vibrations. Although the highest-resolution co-located SEHA array in our survey had 11.1 pixels/cm$^2$ \cite{Leroy_Shea_2023}, Purenendu et al. \cite{purnendu2023fingertip} built a 16 pixels/cm$^2$ array using routed SEHAs (Fig. \ref{fig:lit_review_images_cutaneous}(a.iii)). By placing larger SEHA pumping pouches on the back of the fingertip and connecting them to contact units via elastomeric tubes, each pixel could achieve up to 1.66 mm stroke and 60 mN force output, although the extra tubing distance decreased their actuator's bandwidth to about 7 Hz (Fig. \ref{fig:lit_review_images_cutaneous}(a.iii)). 

Using EOP EKPs, Shen et al. \cite{Shen_Rae-Grant_Mullenbach_Harrison_Shultz_2023} developed Fluid Reality (Fig. \ref{fig:lit_review_images_cutaneous}(a.iv)), a fully self-contained haptic glove featuring 160 individually addressable pixels across all five fingerpads. By operating at 300 V, compared to the kiloVolt ranges typically required by SEHAs, they were able to miniaturize the electronics into the glove's wristband for truly untethered wearable haptic experiences. Each fingerpad had a spatial resolution of 20 pixels/cm$^2$, and pixel could deliver millimeter-scale displacements up to frequencies of hundreds of Hertz for expressive tactile feedback. Through user studies in VR, participants could accurately recognized static and animated shapes presented on an individual fingerpad, and they reported that the glove could help recreate the properties of virtual objects, such as shape, texture, and compliance, through tactile exploration. These sensations were well perceived and reported to be distinguishable, realistic, expressive, and harmonious. Yu et al. \cite{yu2025morphingskin} later extended EOP EKPs into MorphingSkin, a flexible EOP array that can conform to the hand.  %  without requiring a rigid backing.
% small-scale tactile renderings such as 

% With a thickness of only 5 mm, each unit delivers up to 50 kPa peak pressure, 1 mm displacement, and operates within a 0–320 Hz bandwidth. Unlike SEAs, which typically require 2–6 kV to actuate with a bulky power supply, the EOP-based design operates under 300 V and its driver and control system is miniaturized for mounting on the wrist, enabling truly untethered wearable haptic interactions. The paper also demonstrated its capability in VR applications to convey rich haptic features, including complex contact geometries, textures, compliance, and expressive spatiotemporal effects \cite{Shen_Rae-Grant_Mullenbach_Harrison_Shultz_2023}.

% \paragraph{World-Grounded and Handheld Devices}
\subsubsection{\revision{Normal Skin Deformation with Optional Vibrotactile Feedback (World-Grounded and Handheld Devices)}}

OOP HVEAs have also been applied in world-grounded and handheld applications to enrich everyday interactions with objects and surroundings. For instance, Kim et al. \cite{Kim_Nam_Kim_Kyung_2021} (Fig. \ref{fig:lit_review_images_cutaneous}(c.i)) attached a thin, flexible SEHA to a car's steering wheel, providing localized vibration to deliver navigation cues to the driver. In a medical context, Lee et al. \cite{lee2017development} integrated DEAs into a handheld, pen-shaped surgical tool. The tool also included a 3-DoF force sensor on the probe, and its measurements were translated into haptic feedback via three DEA units around the tool handle's circumference.  % for the grasping fingers

% \textit{Refreshable braille displays} present a promising real-world application of HVEAs, and they serve as a common performance benchmark for out-of-plane pixel arrays in our literature review. Braille characters consist of a 3x2 dot matrix (4x2 for Unicode characters), with each dot approximately 1.5 mm in diameter, 0.6 - 0.9 mm tall, and spaced about 2.5 mm apart (corresponding to a density of 16 dots/cm$^2$). Refreshable braille displays must be able to resist normal interaction forces from the user ($\sim$0.2 - 0.5 N \cite{Smith_Gosselin_Houde_2002, Runyan_Blazie_2011}), and they should have refresh rates $>$10 Hz to match normal braille reading speeds. Commercial displays usually feature 40 - 80 braille cells, comprising up to 640 individually addressable pixels, although larger units meant for two-handed reading can have up to 720 braille cells. Refreshable Braille displays on the market are powered by bulky piezoelectric bimorphing beams that can cost many thousands of dollars, making portable Braille displays cost- and transport-prohibitive \cite{Runyan_Blazie_2011}.

\textit{Refreshable braille displays} represent a promising real-world application of HVEAs and a common benchmark for OOP actuator arrays. These devices deliver raised-dot patterns of braille characters according to standardized tactile guidelines \cite{Smith_Gosselin_Houde_2002, Runyan_Blazie_2011}, which require sufficient deformations, resistive forces, and inter-dot spacings for clear fingertip perception. They also need to refresh fast enough to support natural reading speeds. Commercial displays typically include 40-80 braille cells (comprising up to 640 individually addressable dots), while larger units for two-handed reading can have up to 720 braille cells. However, devices on the market mostly rely on bulky piezoelectric bimorphing beams and cost many thousands of dollars, limiting portability and affordability \cite{Runyan_Blazie_2011}.

The manufacturing scalability and high spatial resolution of DEAs present a promising fit for refreshable braille displays, comprising 20 papers in our literature review. The largest fabricated braille display among them is by Ren et al. \cite{Ren_Liu_Lin_Wang_Zhang_2008}, who made 324 braille cells by spray coating all of the electrodes and dielectric onto a substrate using an airbrush and a laser-cut mask. Each of their braille cells consisted of 3x2 DEA dots, totaling 1944 individually addressable pixels. While many planar DEAs can achieve the required deformation at high driven voltages \cite{Niu_Brochu_Stoyanov_Yun_Pei_2012, Qu_Ma_Shi_Li_Zheng_Wang_Liu_Fan_Chen_Li_2021}, 
rolled DEAs typically deliver superior performance. Levard et al. \cite{Levard_Diglio_Lu_Gorny_Rahn_Zhang_2011}, for instance, developed a rolled DEA braille cell that could output displacements up to 1.2 mm and block forces over 1 N, well exceeding the tactile requirement for braille reading.

\textit{Pushbuttons} are another common application, comprising 12 papers in our survey. Many implementations include both OOP actuation and sensing for bidirectional interaction. Chen et al. \cite{chen2022haptag} developed HapTag, which integrates a sensing layer beneath an OOP SEHA unit, creating a soft, compact button that can be attached to daily soft surfaces. Also using SEHAs, Firouzeh et al. \cite{firouzeh2024poptouch} developed PopTouch, a transparent button for high-bandwidth actuation. Their sensing was achieved either through integration into a capacitive touchscreen or by overlaying the actuator with thin-film piezoelectric sensors. PopTouch also used novel anchored-HAXELs, featuring a snap-through force-displacement profile that convincingly replicates the familiar ``click'' of physical buttons. 

\revise{
EKPs have also been used to design dynamic tactile buttons. Fujii et al. \cite{fujii2021layerpump} developed a soft haptic button using stacked EHDPs, as seen in Fig. \ref{fig:lit_review_images_cutaneous}(c.ii). When activated, dielectric liquid is pumped into the chamber, raising the top elastomeric membrane. Rae-Grant et al. designed DynaButtons \cite{Rae-Grant2024dynabuttons}, which integrated an EOP EKP with a pressure-sensing layer for enable fast, expressive, closed-loop button interactions. 
% and layered acrylic plates to precisely guide fluid flow.  to provide a haptic sensation when the user places their finger on it
}

Gratz-Kelly et al. \cite{Gratz-Kelly_Kruger_Seelecke_Rizzello_Moretti_2024} expanded these techniques to develop an audio-tactile button, driving a frame-coupled DEA with low-frequency, high-voltage signals for haptic actuation; high-frequency, low-voltage signals for proprioceptive capacitance sensing; and high-frequency, high-voltage signals to vibrate the DEA and produce sound. They demonstrated that multiple of these drive schemes could be added together to create diverse audio-tactile experiences. 

% \subsubsection{Lateral Skin Stretch with Optional Vibrotactile Feedback}
\subsubsection{\revision{Lateral Skin Stretch with Optional Vibrotactile Feedback (Wearables)}}

Compared to normal forces, human skin is roughly 2–3× more sensitive to lateral (shear) forces \cite{Biggs_Srinivasan_2002}. In daily life, lateral skin stretch commonly occurs passively when fingers slide across surfaces with varying friction, and Sec.~\ref{subsubsec:variable_friction_surfaces} will review how ESAs have been extensively used to create programmable-friction haptic surfaces. In this section, we focus on localized lateral skin stretch actively delivered by HVEA actuators. \revision{8 papers in our literature review explored HVEAs, primarily using DEAs and SEHAs, for wearable applications, and 8 papers explored HVEAs for world-grounded applications.}
% , which 16 papers in our survey explored.  % 16 = 11 + 2 (normal + shear) + 3 (HAXELs in normal)
% passive variable

% \paragraph{Wearable Devices}
% Lateral skin stretch has been implemented in several wearable devices. 
Bolzmacher et al. \cite{bolzmacher2006flexible} presented one of the first wearable HVEAs for lateral skin stretch in 2006 by wrapping a fiber-reinforced DEA around a participant's arm. They report that the DEA's expansion in-plane produced discernable skin stretch, but their device also required an external high voltage amplifier, limiting portability. Ji et al. \cite{Ji_Liu_Cacucciolo_Civet_ElHaitami_Cantin_Perriard_Shea_2021} later introduced untethered feel-through haptics by mounting an ultra-thin (0.018 mm), flexible DEA film and a custom low-profile 480 V power supply onto the fingertip (Fig. \ref{fig:lit_review_images_cutaneous}(d.i)). Their actuator's in-plane expansion can deliver a well-localized pulsing sensation at low frequencies (1–20 Hz), while higher frequencies (100–500 Hz) produce a vibration felt across the entire fingertip. HAXEL SEHAs can also produce shear forces, as noted in Sec.~\ref{subsubsec:normal_skin_deformation}, and generally offer larger forces and displacements but lower spatial resolutions and bandwidths than comparable DEA solutions \cite{Leroy_Shea_2023}.
% the OOP SEA actuators that normally deliver normal skin deformation and vibration, can also generate shear forces when operated with split electrodes.
% , with or without a flexible frame  When a voltage is applied, the thin DEA expands in-plane, which gently stretches the skin. 

\subsubsection{\revision{Lateral Skin Stretch with Optional Vibrotactile Feedback (World-Grounded Devices)}}
% \paragraph{World-grounded Devices}
% 8 papers in our literature review applied HVEAs to render shear cutaneous feedback in world-grounded applications, primarily using DEAs. 
Han et al. \cite{Han_Bae_Gregoriou_Ploch_Goldman_Glover_Daniel_Cutkosky_2018} used DEAs to deliver localized lateral skin stretch to the thumb and index fingertips of an MRI-compliant graspable operator, conveying forces sensed by a robotized biopsy needle (Fig. \ref{fig:lit_review_images_cutaneous}(d.ii)). Notably, HVEAs can be built without metallic components, ensuring compatibility for MRIs and in other magnetic environments. DEA arrays have also been explored for rendering spatial skin stretch patterns; for example, Knoop et al. \cite{Knoop_Rossiter_2014} developed a 5x5 hinged pin array, which used a continuous DEA membrane underneath the pins to deliver controlled lateral skin stretch across a fingertip.
% \cite{yamamoto2005evaluation}
% DEA in-plane expansion has also been explored for pixel arrays to deliver lateral skin stretch. 
% As the DEA has patterned electrodes, selective actuation of an element of the DEA causes localized expansion, moving the adjacent pins laterally.
% knoop2015compliant

\subsubsection{Miscellaneous Cutaneous Feedback Interactions} \label{subsubsec:miscellaneous_cutaneous_feedback}
HVEAs can also deliver cutaneous feedback beyond skin deformation. For instance, Iriarte et al. \cite{Iriarte_Ezcurdia_Elizondo_Irisarri_Hemmerling_Ortiz_Marzo_2024}, shown in Fig. \ref{fig:lit_review_images_cutaneous}(e.i), and Fukushima and Kajimoto \cite{Fukushima_Kajimoto_2012} demonstrated contactless electrostatic piloerection by charging electrodes up to 5-20 kV and positioning them a few centimeters away from participants’ napes, forearms, and hands. The resulting electrostatic attraction lifted participants' body hair, creating goosebumps. Most participants described the stimulation as comfortable and pleasant after the initial surprise, although both studies also reported large individual differences in piloerection sensitivity.

% Beyond mechanoreceptors, nocireceptors, also known as pain receptors, and thermoreceptors are free nerve endings that occupy much of the same areas of the skin's dermis and epidermis as mechanoreceptors, and thus many taxonomies also include them as a type of cutaneous feedback \cite{lederman2009haptic, goldstein2002sensation}. 
%HVEAs have also been applied to stimulate other receptors in the skin, such as nocireceptors (pain receptors) and thermoreceptors. 
% Mujibiya \cite{Mujibiya_2015} explored nociceptive feedback by injecting charges into users' feet to trigger a controllable-intensity electric shock whenever the user touched a grounded surface. 
Electrohydrodynamic EKPs have also been employed for thermal stimulation, with five out of the six EHDPs in our literature review implementing fiber-based pumps that can transport hot or cold liquid from a reservoir through tubes wrapped around a user's hand or body. The highest flow rate EHDP, shown in Fig. \ref{fig:lit_review_images_cutaneous}(f.i), could pump up to 55 mL/min, allowing cold liquid to move from a reservoir on the back of a user's hand through tubes across the entire hand within two seconds, or from a reservoir on a user's chest through tubes across their entire torso within five seconds \cite{Smith_Cacucciolo_Shea_2023}.

% \begin{figure*}[th]
%     \centering
%     \includegraphics[width=\linewidth]{figures/kinesthetic_feedback.pdf}
%     % \caption{Examples of HVEA haptic applications targeting kinesthetic feedback. The images were drawn from (a.i) \cite{Hinchet_Shea_2022} (\copyright 2022 The Authors. Advanced Intelligent Systems published by Wiley-VCH GmbH), (b.i) \cite{Ramachandran_Schilling_Wu_Floreano_2021} (\copyright 2021 The Authors. Advanced Intelligent Systems published by Wiley-VCH GmbH), (b.ii) \cite{Feizi_Atashzar_Kermani_Patel_2022} (\copyright 2022 IEEE), (c.i) \cite{Diller_Majidi_Collins_2016} (\copyright 2016 IEEE), (d.i) \cite{Hinchet_Shea_2022} (\copyright 2022 The Authors. Advanced Intelligent Systems published by Wiley-VCH GmbH), (e.i) \cite{Hara_Matthey_Yamamoto_Chapuis_Gassert_Bleuler_Higuchi_2009} (\copyright 2009 IEEE).}

%     \caption{Examples of HVEA haptic applications targeting kinesthetic feedback. The images were drawn from (a.i \& a.ii \& a.iv) \cite{Hinchet_Shea_2022} (\copyright 2022 The Authors. Advanced Intelligent Systems published by Wiley-VCH GmbH), (a.iii) \cite{Ramachandran_Schilling_Wu_Floreano_2021} (\copyright 2021 The Authors. Advanced Intelligent Systems published by Wiley-VCH GmbH), (a.v) \cite{Diller_Majidi_Collins_2016} (\copyright 2016 IEEE), (b.i) \cite{Feizi_Atashzar_Kermani_Patel_2022} (\copyright 2022 IEEE), (b.ii) \cite{Hara_Matthey_Yamamoto_Chapuis_Gassert_Bleuler_Higuchi_2009} (\copyright 2009 IEEE).}
%     \label{fig:lit_review_images_kinesthetic}
% \end{figure*}

\begin{figure}[t]
    \centering
    \includegraphics[width=\linewidth]{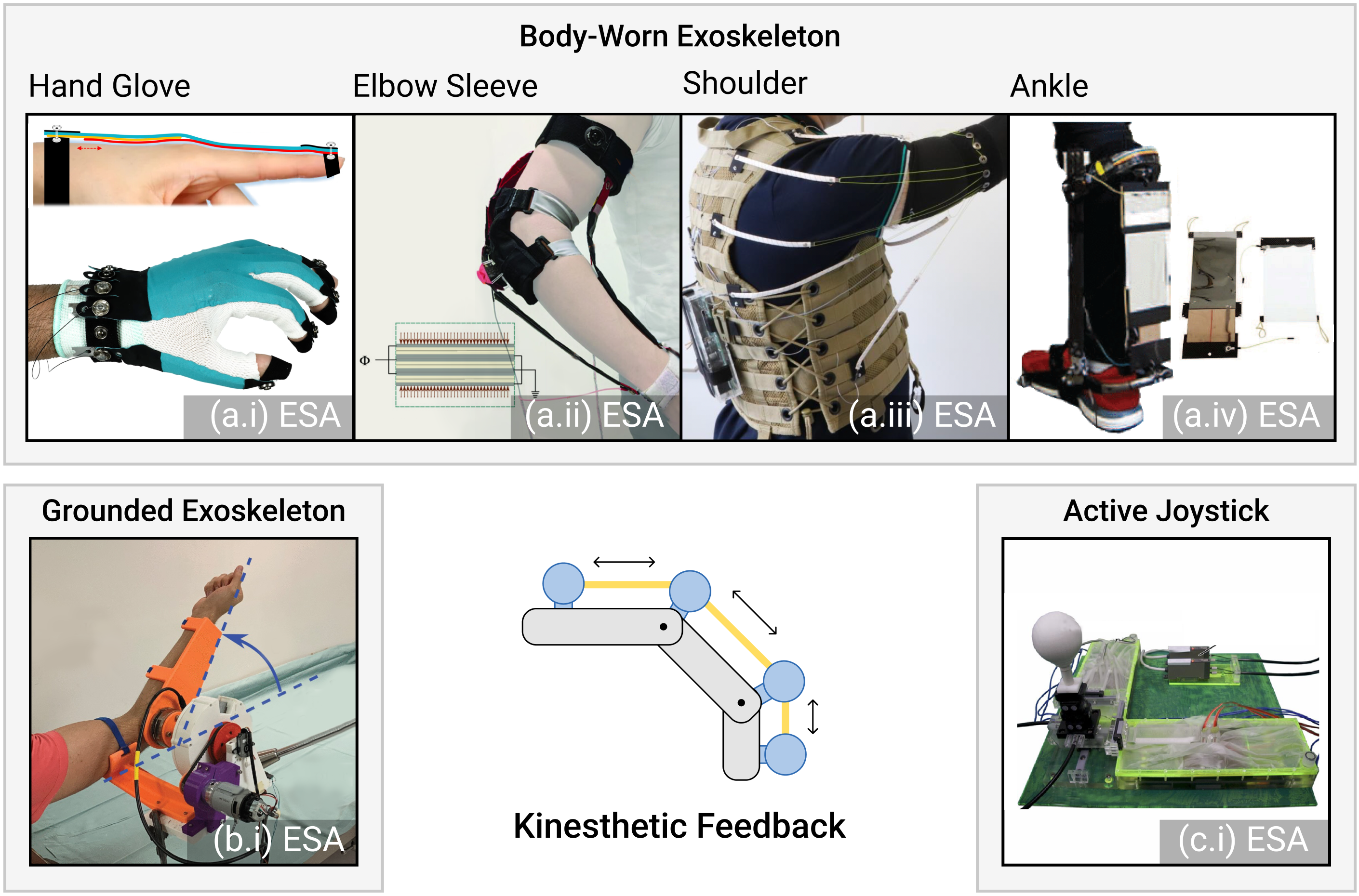}
    % \caption{Examples of HVEA haptic applications targeting kinesthetic feedback. The images were drawn from (a.i) \cite{Hinchet_Shea_2022} (\copyright 2022 The Authors. Advanced Intelligent Systems published by Wiley-VCH GmbH), (b.i) \cite{Ramachandran_Schilling_Wu_Floreano_2021} (\copyright 2021 The Authors. Advanced Intelligent Systems published by Wiley-VCH GmbH), (b.ii) \cite{Feizi_Atashzar_Kermani_Patel_2022} (\copyright 2022 IEEE), (c.i) \cite{Diller_Majidi_Collins_2016} (\copyright 2016 IEEE), (d.i) \cite{Hinchet_Shea_2022} (\copyright 2022 The Authors. Advanced Intelligent Systems published by Wiley-VCH GmbH), (e.i) \cite{Hara_Matthey_Yamamoto_Chapuis_Gassert_Bleuler_Higuchi_2009} (\copyright 2009 IEEE).}

    \caption{Examples of HVEA haptic applications targeting kinesthetic feedback. The images were drawn from (a.i \& a.iii) \cite{Hinchet_Shea_2022} (\copyright 2022 The Authors. Advanced Intelligent Systems published by Wiley-VCH GmbH), (a.ii) \cite{Ramachandran_Schilling_Wu_Floreano_2021} (\copyright 2021 The Authors. Advanced Intelligent Systems published by Wiley-VCH GmbH), (a.iv) \cite{Diller_Majidi_Collins_2016} (\copyright 2016 IEEE), (b.i) \cite{Feizi_Atashzar_Kermani_Patel_2022} (\copyright 2022 IEEE), (c.i) \cite{Hara_Matthey_Yamamoto_Chapuis_Gassert_Bleuler_Higuchi_2009} (\copyright 2009 IEEE).}
    \label{fig:lit_review_images_kinesthetic}
\end{figure}

\subsection{Kinesthetic Feedback} \label{subsec:kinesthetic_feedback}
\revise{Kinesthetic feedback conveys a sense of force, motion, or resistance through muscles and joints. Among the reviewed studies, only ESAs were employed to generate this form of feedback, using electrostatic attraction to both passively brake and actively drive movement.}
% Kinesthesia is defined as the body's sense of movement and position, and kinesthetic feedback is produced by applying forces and torques that can be sensed in a user's muscles, tendons, and joints. 
% ESAs were the only HVEAs used for kinesthetic feedback, but in Sec. \ref{sec:future_work} we discuss future pathways to achieve kinesthetic feedback using other HVEAs. 

\subsubsection{Passive Kinesthetic Feedback} \label{subsubsec:passive_kinesthetic_feedback}
As shown in Fig. \ref{fig:lit_review_images_kinesthetic}(a, b), electrostatic switchable adhesives can be integrated into clothing or exoskeletons to programmatically brake and dissipate energy during a person's motion. These ESAs comprise two conductive strips, either thin metal shims or conductive fabric strips, coated in a thin dielectric and mounted to different bony protrusions in the body. When the voltage is off, the two conductive strips can deform with the user's body and slide against each other. When the two strips are held at a potential difference, however, they adhere together and resist further movement by the user. We found 7 papers integrating ESAs into haptic gloves, 6 on haptic sleeves, 3 on shoulder exoskeletons, and 1 each on ankle and back exoskeletons. 

Gloves were the most common form factor of passive kinesthetic feedback in our survey. Every paper mentioned VR as the primary use case, taking advantage of many VR headsets' integrated hand tracking. Hands can move in complex ways, and choosing the right locations to mount ESAs on the hand is particularly important to deal with metacarpal abduction (particularly for the thumb). Hinchet et al. \cite{Hinchet_Vechev_Shea_Hilliges_2018} found that mounting one conductive strip to the first joint of each finger and the second conductive strip to the wrist works well, with the thumb's clutch tilted 30$^\circ$ outwards to help cope with its more complex flexion and abduction motion. The strips also need to be wrapped in either a fabric or plastic guide sheath to ensure they only deform along each finger's axis. Hinchet and Shea \cite{Hinchet_Shea_2022} developed our survey's highest performing glove, shown in Fig. \ref{fig:lit_review_images_kinesthetic}(a.i), which could block up to 50 N per finger at only 300 V. The fabric construction meant the haptic glove was thin (1.3 mm), lightweight (28 g), and fast ($<$ 100 ms response time). It also conformed well to users' hands, enabling accurate finger tracking for VR applications.
% used 125 \textmu m thick metallized mylar strips coated in a 25 \textmu m layer of P(VDF-TrFE-CTFE)
% 7 of the 8 studies also performed a virtual reality (VR) user study, using VR headsets with integrated hand tracking and measuring whether the haptic gloves 

% Haptic sleeves were the next most common form factor of passive kinesthetic feedback explored
ESAs have also been implemented in sleeves and arm exoskeletons, where clutches are positioned between the forearm and upper arm on either the upper arm's ventral side (to restrict elbow extension) or its dorsal side through a guide on the elbow joint (to restrict elbow flexion). The best performing haptic sleeves were by Zhang et al. \cite{Zhang_Shea_2025}, who engaged three clutches in parallel at 300 V to block up to 304.8 N, and Ramachandran et al. \cite{Ramachandran_Schilling_Wu_Floreano_2021} (shown in Fig. \ref{fig:lit_review_images_kinesthetic}(a.ii)), who made a single clutch that could block up to 92.4 N at 400 V. These results compare favorably against human performance studies, which show most people cannot produce more than 116 N of static force during elbow flexion and extension \cite{Human_Engineering_2020}, although Hinchet and Shea \cite{Hinchet_Shea_2022} showed that even 5 N is sufficient for many VR discrimination tasks. Beyond VR applications \cite{Vechev_Hinchet_Coros_Thomaszewski_Hilliges_2022, Hinchet_Shea_2022}, haptic sleeves have been explored for drone teleoperation \cite{Ramachandran_Macchini_Floreano_2022} and motor learning \cite{Ramachandran_Schilling_Wu_Floreano_2021}, while grounded arm exoskeletons have supported rehabilitation usecases\cite{Feizi_Atashzar_Kermani_Patel_2022} (Fig. \ref{fig:lit_review_images_kinesthetic}(b.i)).
% , and haptic feedback while in virtual reality \cite{Vechev_Hinchet_Coros_Thomaszewski_Hilliges_2022,Hinchet_Shea_2022}.
% citations for restricting extension on the ventral side/dorsal side for elbow extension =  \cite{Ramachandran_Schilling_Wu_Floreano_2021, Hinchet_Shea_2022, Ramachandran_Macchini_Floreano_2022, Ramachandran_Shintake_Floreano_2019}

Other body-worn ESA clutches have been explored in various exoskeleton form factors for improving movement efficiency and reducing muscle exertion. Diller et al. \cite{Diller_Collins_Majidi_2018} mounted five ESA clutches in parallel between a person's knee and shoe to serve as an ankle exoskeleton, as shown in Fig. \ref{fig:lit_review_images_kinesthetic}(a.iv). When clutched, the ESAs adhered to a rubber spring that would then stretch and provide passive kinesthetic feedback as the person's ankle flexed during their gait. The exoskeleton could block up to 501 N with response times as fast as 30 ms, and during a person's gait it could exert up to 7.3 N$\cdot$m on an average step. Hinchet and Shea \cite{Hinchet_Shea_2022}, as shown in Fig. \ref{fig:lit_review_images_kinesthetic}(a.iii), integrated eight long, narrow ESA clutches around the shoulder that could block different arm movements. Each actuator could reportedly block up to 196 N. These results compare favorably even against human performance studies, which show most people cannot produce more than 250 N of static force during shoulder flexion and extension \cite{Human_Engineering_2020}.

% Vechev et al. \cite{Vechev_Hinchet_Coros_Thomaszewski_Hilliges_2022} integrated ESAs into textiles to create a programmable stiffness vest. Hinchet et al. \cite{Hinchet_Shea_2022, Hinchet_Vechev_Shea_Hilliges_2018} coated strips of metallized PET with a thin P(VDF-TrFE-CTFE) dielectric to form individual ESA clutches for each fingertip and the elbow. Diller et al. \cite{Diller_Majidi_Collins_2016} combined ESAs with a rubber spring to provide passive kinesthetic feedback during a person's gait. Mention series-elastic actuators (SEAs).

Beyond exoskeletons, ESAs have also been attached on objects to provide passive kinesthetic feedback. Mishra et al. \cite{Mishra_Kumar_Shukla_Parnami_2022} integrated ESAs with co-planar interdigitated electrodes on the bottom of everyday objects like a computer mouse, cardboard box, and fish toy to produce up to 5 N of kinetic friction given an input voltage of 2.3 kV when participants dragged the objects across a flat tabletop.

\subsubsection{Active Kinesthetic Feedback} \label{subsubsec:active_kinesthetic_feedback}
A specific form factor of ESAs, called ``Dual Excitation Multiphase Electrostatic Drive'' (DEMED), can also be used to generate active forces to drive and guide user movement rather than only resisting it. DEMEDs etch a three-phase skewed electrode pattern into both sides of an ESA, and when both sides are driven with a three-phase input signal (the top electrodes are phase shifted by 120$^\circ$ relative to the bottom electrodes), the 120$^\circ$ phase shift and skewed electrode pattern result in a traveling wave of shear force pushing both sides apart. Multiple DEMEDs can also be stacked in parallel to amplify the output force. We found two papers by Hara et al. \cite{Hara_Matthey_Yamamoto_Chapuis_Gassert_Bleuler_Higuchi_2009} and Kimura et al. \cite{Kimura_Yamamoto_Hara_Ryu_Bleuler_Higuchi_2011} integrating eight-layer DEMEDs into haptic joysticks. Hara et al.'s haptic joystick, shown in Fig. \ref{fig:lit_review_images_kinesthetic}(c.i), could produce up to 18.0 N of force and 3.1g accelerations using a 2.4 kV, 10 Hz drive signal. Kimura et al.'s haptic joystick demonstrated up to 7 N of force using a 1.4 kV, 330 Hz drive signal.

\subsection{Haptic Surfaces} \label{subsec:explorable_hapitc_surfaces}
% We categorize shape displays as world-grounded kinesthetic devices, in line with prior haptics taxonomies \cite{culbertson2018haptics}. Shape displays are large actuator arrays that can rise up out of a grounded state, say a tabletop, to render an approximation of a 3D surface. They allow for whole-hand haptic interactions, enabling tactile manipulation of CAD models and accessible 3D prototyping tools for blind or visually impaired makers. Shape displays can also be used for art or aesthetic purposes, where shapes can pop out to augment a visual experience \cite{Alexander_Roudaut_Steimle_Hornbaek_Bruns_Alonso_Follmer_Merritt_2018}.

% Another perspective on the creation of haptic experiences is that they could be real-world surfaces that users can actively explore, enabling simultaneous kinesthetic and cutaneous feedback. Such haptic surfaces would ideally feel just like natural objects in the environment, except that they can arbitrarily change their shape, mechanical properties, and surface texture. 

\revise{Haptic surfaces represent another major class of interfaces that allow users to actively explore with their hands as the surface dynamically alters its texture or shape. This category includes surfaces that vary their friction as the fingers slide across them, as well as those that integrate near-silent, high-performance vibrotactile actuators. HVEAs have also been implemented in tabletop pin arrays and deformable crust surfaces that render dynamic shape change, providing simultaneous kinesthetic and cutaneous feedback when users interact with them freely with their hands.} 
Haptic surfaces typically integrate sensing to respond dynamically to user interaction. To properly render desired shapes, the actuators need to be able to withstand the roughly 0.5 - 1 N normal forces applied by typical users during fingertip exploration \cite{Smith_Gosselin_Houde_2002, Pawluk_Howe_1999}. HVEAs' high bandwidth and force density are well suited for creating responsive 2D and 2.5D haptic surfaces. 
% Fig. \ref{fig:lit_review_images_hapticsurface} highlights several examples of previous HVEA haptic surfaces.
% Haptic surfaces represent another major class of interfaces that allow users to actively explore with their hands, while the surface dynamically alters its texture or shape.
% To achieve perceivable shape change

% fast, high degree-of-freedom 2D and 2.5D haptic surfaces. 
% Fig. \ref{fig:lit_review_images_hapticsurface} highlights several examples of previous HVEA haptic surfaces.

\subsubsection{Variable Friction Surfaces} \label{subsubsec:variable_friction_surfaces}

\paragraph{Electrovibration} 
The single most common application in our survey (comprising 83 papers) was integrating ESAs into flat surfaces to create a variable-friction haptic plane. Charging a conductive electrode below a thin dielectric induces an opposite charge in a user's fingertip, creating friction as the user drags their finger along the display. This mechanism, also known as electrovibration, was first proposed in 1970 by Strong and Troxel \cite{Strong_Troxel_1970}. Meyer et al. proposed a carrier wave modulation strategy in 2014 that takes advantage of the unique electric double layer properties of human skin, increasing the frictional bandwidth of these displays \revise{well into the kHz} regime \cite{Meyer_Wiertlewski_Peshkin_Colgate_2014}, and Shultz et al. demonstrated in 2018 that this modulation strategy could even be used to produce audio tones and even music as a user's finger catches and slips against the display's surface at high frequency \cite{Shultz_Peshkin_Colgate_2018}. 
% ESA frictional displays are most commonly implemented using the unipolar form factor in Fig. \ref{fig:hvea_principles_mechanisms} with a user's fingertip as the substrate.

\begin{figure}[t]
    \centering
    \includegraphics[width=\linewidth]{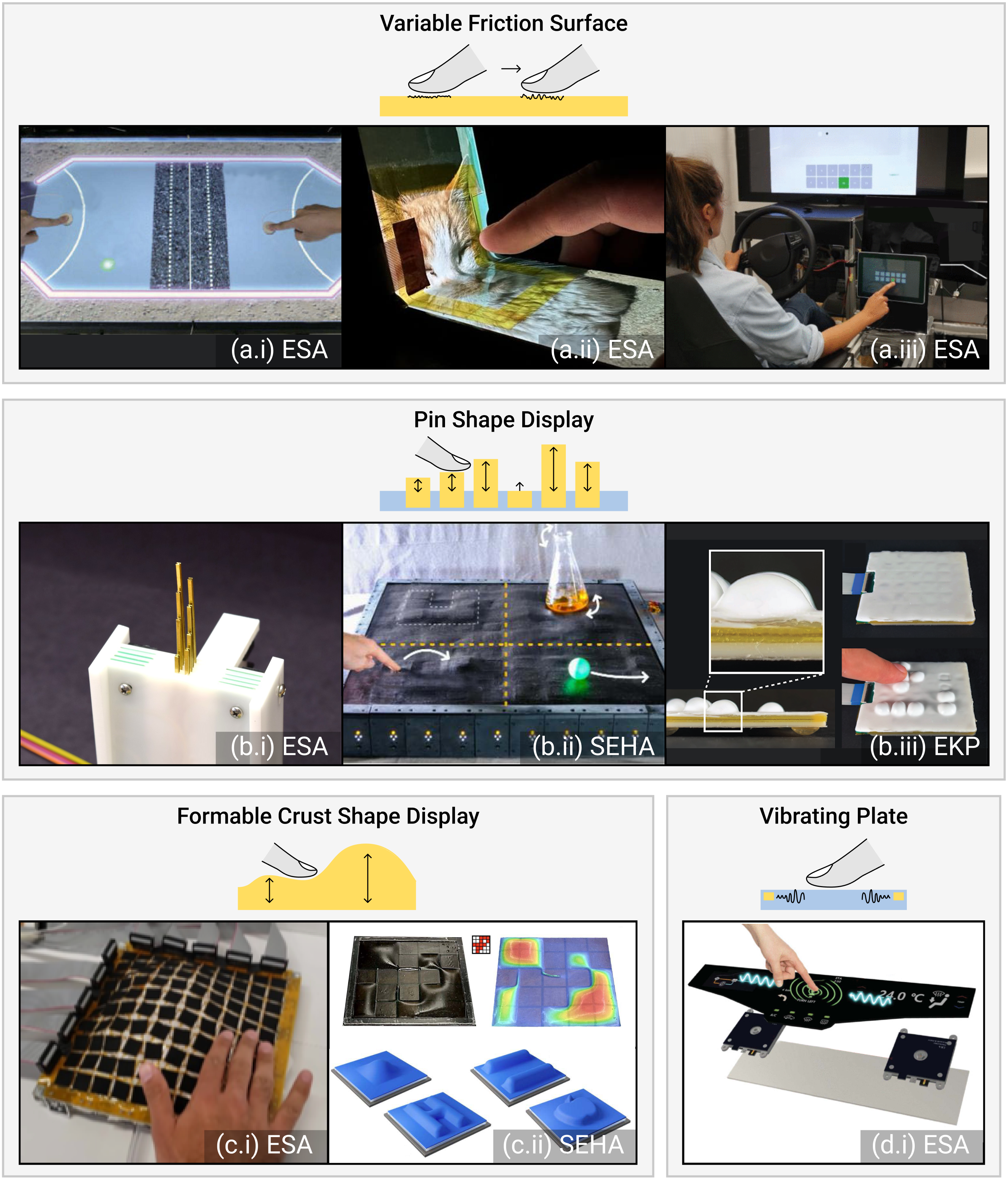}
    \caption{Examples of HVEA haptic applications targeting haptic surfaces. The images were drawn from (a.i) \cite{Nakamura_Yamamoto_2016} (\copyright 2016 IEEE), (a.ii) \cite{Park_Choi_Byun_Choe_Hong_Yang_Lim_Lee_Jung_2024} (\copyright 2024 American Chemical Society) (a.iii) \cite{Breitschaft_Pastukhov_Carbon_2022} (\copyright 2022 IEEE), (b.i) \cite{Zhang_Gonzalez_Guo_Follmer_2019} (\copyright 2019 IEEE), (b.ii) \cite{Johnson_Naris_Sundaram_Volchko_Ly_Mitchell_Acome_Kellaris_Keplinger_Correll_2023} (\copyright 2023 The Authors), (b.iii) \cite{Shultz_Harrison_2023} (\copyright 2023 The Authors), (c.i) \cite{Rauf_Bernardo_Follmer_2023} (\copyright 2023 IEEE), and (c.ii) \cite{jang2024dynamically} (\copyright 2024 The American Association for the Advancement of Science), (d.i) \cite{Rajkumar_Singh_Yang_Koo_2023} (\copyright 2023 The Authors).}
    \label{fig:lit_review_images_hapticsurface}
\end{figure}

% The most common explorable haptic surface in our literature review was ESAs integrated into displays to create a variable-friction touchscreen. This application works using the electrovibration or reverse electrovibration mechanisms discussed in Sec. \ref{subsubsec:vibrotactile_lateral_skin_stretch}, except that 
Electrovibration can also be integrated into touchscreens, where the display's surface serves as a dielectric covering a transparent electrode. The electrode is typically made of indium-thin oxide (ITO) given its low cost and moderate bend resistance, but flexible touchscreens that expect frequent bending (for example, for folding smartphones, as shown in Fig. \ref{fig:lit_review_images_hapticsurface}(a.ii)) also have explored soft electrodes made from conductive polymers \cite{Park_Choi_Byun_Choe_Hong_Yang_Lim_Lee_Jung_2024}. The idea of an electrovibration touchscreen was first proposed by Yamamoto et al. in 2003 \cite{Yamamoto_Ishii_Higuchi_2003}, and it gained more widespread application after 2010 \revise{by leveraging commercial platforms such as the Senseg Feelscreen \cite{Linjama_Makinen_2009}, 3M SCT3250 MicroTouch  \cite{Bau_Poupyrev_Israr_Harrison_2010}, and TanvasTouch \cite{tanvastouch}}. 
% materials such as graphene sheets \cite{Radivojevic_Beecher_Bower_Haque_Andrew_Hasan_Bonaccorso_Ferrari_Henson_2012}

ESA frictional touchscreens are most commonly implemented by directly using a user's fingertip as the substrate (31 papers), as illustrated in Fig. \ref{fig:hvea_principles_mechanisms}(a.v), but some studies have also explored attaching grounded electrodes to the user's fingertips or having the user hold a grounded stylus (6 papers). Multi-touch interactions can be enabled by adding multiple electrodes underneath the display surface, in the former setup \cite{Haga_Sugimoto_Yang_Sasaki_Asai_Shigemura_2019, Ilkhani_Samur_2018}, or by adding extra grounded contact pads for each finger in the latter case \cite{Nakamura_Yamamoto_2016, Nakamura_Yamamoto_2013}. The highest-resolution multi-electrode setup in our survey, developed by Haga et al. \cite{Haga_Yoshinaga_Yanase_Sugimoto_Takatori_Asada_2014}, comprises a 52x32 grid of electrodes underneath a 4.1'' diagonal screen. The array's 1.75 mm pitch (66.1 electrodes/cm$^2$) is much finer than that of traditional Braille displays (16 dots/cm$^2$), and modulation strategies can further improve this spatial resolution. For example, Yamamoto et al. \cite{Yamamoto_Ishii_Higuchi_2003} found that scaling the drive voltage frequency with the user's fingertip speed allowed their display to replicate surface textures roughly 2x finer than its physical electrode pitch.

Electrovibration touchscreens can be much quieter than traditional haptic touchscreens based on conventional linear resonant actuators (LRAs) or eccentric rotary mass (ERM) actuators. Prior work has taken advantage of this near-silence for methodical activities such as digital painting \cite{Kim_Osgouei_Choi_2017}, or by adding separate audio overlays for multi-modal haptics in automotive user interfaces \cite{Breitschaft_Pastukhov_Carbon_2022} (Fig. \ref{fig:lit_review_images_hapticsurface}(a.iii)) and interactive tools for people with visual impairments \cite{Feitl_Kreimeier_Gotzelmann_2022, Yoo_Lim_Cho_Choi_2019}. Multi-electrode setups can further enable novel interactions not possible with traditional technologies, including multi-player virtual hockey games with personalized friction feedback, as shown in Fig. \ref{fig:lit_review_images_hapticsurface}(a.i) \cite{Nakamura_Yamamoto_2016}, and two-handed exploration of a dynamically textured surface \cite{Haga_Asai_Takeuchi_Sasaki_Yamamoto_Shigemura_2021}.
% , as shown in Fig. \ref{fig:lit_review_images_hapticsurface}(a.iv) \cite{Haga_Asai_Takeuchi_Sasaki_Yamamoto_Shigemura_2021}. 
 % normal vibrotactile actuators on the back of displays (still ESAs) \cite{Rajkumar_Singh_Yang_Koo_2023, Smith_Gorlewicz_2017}

Integrating ESAs into touchscreens have been shown in many studies to offer performance and qualitative improvements over purely visual displays \cite{Liu_Lv_Wang_Sun_2021, Yan_Li_Sun_Liu_2019, Wang_Sun_Cao_Liu_2022, Zhang_Harrison_2015, Kim_Osgouei_Choi_2017, Liu_Sun_Wang_Liu_Zhang_2018} and traditional vibrotactile displays \cite{Smith_Gorlewicz_2017} for sighted participants. However, for users with visual impairments, preliminary user studies using single-electrode ESA touchscreens show they lag behind traditional dot matrix displays in both performance and qualitative metrics \cite{Feitl_Kreimeier_Gotzelmann_2022, Bateman_Zhao_Bajcsy_Jennings_Toth_Cohen_Horton_Khattar_Kuo_Lee_2018}. Further testing should explore the effectiveness of multi-electrode touchscreens for multi-finger and two-handed interaction.

% Electrovibration can also be coupled with speakers \cite{Breitschaft_Pastukhov_Carbon_2022, Feitl_Kreimeier_Gotzelmann_2022} or ferrofluids (which change viscosity in response to a magnetic field) \cite{Hashizume_Takazawa_Koike_Ochiai_2017} for effective multi-modal haptics.
% An interesting study by Nakamura and Yamamoto \cite{Nakamura_Yamamoto_2014} had users interact with a grounded 40'' LCD monitor while wearing gloves with conductive contact pads on the index and thumb fingertips. The voltage on each contact pad was toggled dynamically as the user interacted with objects on-screen, and each pad could produce up to 1 N of static friction given an input voltage of 400 V. Unlike with traditional passive haptic systems, which create a ``sticky wall'' effect because the user still feels a force when moving away from a boundary, ESA's fast release speeds reportedly allowed for more immersive interaction. ESAs
% While interesting from a haptics perspective, further research still needs to be conducted to how best to integrate electrostatic interactions into touchscreens. Side by side comparisons generally show that electrostatic haptic feedback can improve targeting speed and reduce tracking error compared to conventional flat touchscreens \cite{Zhang_Harrison_2015, Kim_Osgouei_Choi_2017}. Qualitative user studies also tend to show that users find electrostatic haptic feedback more pleasant to use and responsive than a traditional touchscreen \cite{Kim_Osgouei_Choi_2017}. However, \cite{Bateman_Zhao_Bajcsy_Jennings_Toth_Cohen_Horton_Khattar_Kuo_Lee_2018}

\paragraph{Reverse Electrovibration}
As opposed to electrovibration ESAs, which ground the user's body and apply charge on a surface, ``reverse electrovibration'' ESAs apply a charge to the user's body which allows them to induce an opposite charge in any generic object in the field connected to earth ground. Bau and Poupyrev's REVEL interface \cite{Bau_Poupyrev_2012} implement this by strapping a current source to the user's body or to contacting surfaces such as their chair or shoe. The current source injects an AC charge into the user, and the authors reported that users could interact with arbitrary objects in their environment and feel texture even on physically smooth surfaces. The texture depended on the input charge waveform, and they showed that these induced textures could be even coupled with an AR visual interface for perceptual illusions.

Wang et al. \cite{Wang_Ren_Sun_2016} also explored reverse electrovibration to touchscreen interaction by charging a stylus instead of the user's body. This is a much simpler way to achieve multi-point feedback than multi-electrode electrovibration touchscreens, although it requires that each stylus is wired to the touchscreen to equalize their ground potential. 

\subsubsection{Vibrating Plates} \label{subsubsec:vibrotactile_touchscreens}
Eight ESA papers and one DEA paper in our survey papers explored integrating traditional touchscreens with out-of-plane vibrotactile actuators (Fig. \ref{fig:lit_review_images_hapticsurface}(d.i)). The ESA papers called their devices electrostatic resonant actuators (ERAs), inspired by electromagnetic linear resonant actuators. ERAs comprise two large planar electrodes, with the top suspended over the bottom by small spacers or a low-profile spring, and the surfaces vibrate when when a sinusoidal voltage is applied between the two plates. These actuators output the highest deformation at their resonance frequency, and their geometry can be tuned to set this resonance frequency near the optimal sensitivity of skin mechanoreceptors. Mason et al. \cite{mason2020experimental} developed the best-performing ERA in our survey, mounting two 6 x 5.3 cm$^2$ ERAs on either side of a 28 cm-long touchbar. Together, the ERAs could drive a 100 g payload at accelerations up to 8.8g and with a resonance frequency of 196 Hz. The ERAs could also be driven at different frequencies to control the resonance profile along the touchbar's length. 

Yun et al. \cite{Yun_Park_Park_Ryu_Jeong_Kyung_2020} fabricated a large array of fin-like transparent DEAs on a flexible touchscreen panel. Each DEA was 7 x 0.2 x 0.4 mm$^3$ in size, and each and could deform up to 4.1 \textmu m out of plane while producing forces up to 14 mN, enabling localized vibrotactile feedback over the display's length.

\subsubsection{Pin Arrays}
\label{subsubsec:pin_array}
Pin arrays are 2.5D displays that can render a pixelated version of a continuous surface. They use an array of pins that can be independently raised or lowered from a grounded state to a desired height. Pin arrays allow for whole-hand haptic interactions, enabling tactile manipulation of CAD models and accessible 3D prototyping tools for blind or visually impaired makers \cite{Siu_Kim_Miele_Follmer_2019}. They are traditionally made using arrays of commercial linear actuators, which are fast and reliable but are also relatively bulky and power-intensive \cite{Alexander_Roudaut_Steimle_Hornbaek_Bruns_Alonso_Follmer_Merritt_2018}. We identified three studies in our literature review that explored using HVEAs to produce pin arrays.

% Zhang et al. \cite{Zhang_Gonzalez_Guo_Follmer_2019}, depicted in Fig. \ref{fig:lit_review_images_hapticsurface}(b.i), is the most directly analogous to a traditional pin array. They

Most similar to the form of traditional pin arrays, Zhang et al. \cite{Zhang_Gonzalez_Guo_Follmer_2019} fabricated an ESA clutch array with co-planar electrodes, which allowed them to concentrate all of the driving electronics into the pin array housing (Fig. \ref{fig:lit_review_images_hapticsurface}(b.i)). To render a shape, they first used a stepper motor to raise a 4x2 array of brass pins up to a maximum height of 50 mm. They then lowered all of the pins at a constant speed, and when each pin reached its desired height they adhered the pin in place with its corresponding ESA clutch. The system combines a large stroke length, characteristic of traditional pin arrays, with a spatial resolution (34.6 pins/cm$^2$) comparable to that of the best ESA OOP dot array in our review (44.4 pixels/cm$^2$ \cite{Takagi_Sasaki_Shikida_Sato_2007}). Its spatial resolution is also much larger than that of, to the authors' knowledge, the highest resolution DC motor-based pin array with comparable stroke (2.0 pins/cm$^2$) \cite{Siu_Gonzalez_Yuan_Ginsberg_Follmer_2018}. 
% While the 0.35 Hz refresh rate is relatively slow due to the stepper motor's slow speed, the ESA clutch array enables 

% Although fluidic actuators are not traditionally considered pins, we consider the next two studies to serve the same functional purpose as a traditional pin array in rendering 2.5D shapes
SEHAs and EKPs have also been explored as linear actuators for haptic pin arrays, trading off the smaller stroke lengths of their fluidic actuators in exchange for having higher refresh rates. Johnson et al. \cite{Johnson_Naris_Sundaram_Volchko_Ly_Mitchell_Acome_Kellaris_Keplinger_Correll_2023} developed a 10x10 array of stacked SEHA HASEL actuators capable of transmitting forces up to 2.5 N and generating displacements up to 14 mm at 10 Hz refresh rates (Fig. \ref{fig:lit_review_images_hapticsurface}(b.ii)). Each actuator was 6x6x9 cm$^3$, and the authors demonstrated the display's ability for dynamic 2.5D pattern rendering, whole-hand haptic interaction, and moving small objects around the display surface. Shultz and Harrison \cite{Shultz_Harrison_2023} developed a 32-pixel array of EKP electro-osmotic pumps capable of blocking forces up to 1.4 N and generating displacements up to 5.75 mm at lower frequencies (0.25 - 1 Hz). As shown in Fig. \ref{fig:lit_review_images_hapticsurface}(b.iii), each cell was about 1x1x0.5 cm$^3$ and could programmably move between a raised position, neutral, and a lowered position by controlling its electrode voltage polarity. The authors demonstrated the display's ability for dynamic pattern rendering and fingertip haptic interaction. They also demonstrated that the EKP array could be placed underneath a flexible AMOLED screen to couple visual display elements with tactile bumps and vibration.
% Closed-loop position control enabled each cell to have a programmable height. 

\subsubsection{Deformable Crust Surfaces}
Deformable crust shape displays differ from pin arrays by embedding actuation or variable stiffness mechanisms into the surface itself, reducing assembly bulk, facilitating curved surface rendering, and enabling whole-hand surface interaction. These displays and their vision of ``digital clay'' prototyping tools have received much interest in haptics research \cite{Wang_Chortos_2022}. However, traditional variable stiffness mechanisms such as pneumatic layer jamming are often difficult to integrate into thin, elastic surfaces \cite{Steltz_Mozeika_Rodenberg_Brown_Jaeger_2009}, and emerging technologies such as ionic electro-actuated polymers are often fabrication-limited to fingertip-sized displays \cite{Wang_Sotzing_Lee_Chortos_2023}.
% Rossignac_Allen_Book_Glezer_Ebert_Uphoff_Shaw_Rosen_Askins_Bai_Bosscher_2003

Two papers in our review applied HVEAs to produce fast-actuating, high degree-of-freedom deformable crust surfaces. Rauf et al. \cite{Rauf_Bernardo_Follmer_2023}, shown in Fig. \ref{fig:lit_review_images_hapticsurface}(c.i), explored this design space by using electroadhesive auxetic skins to create programmable shape change when wrapped as a strain-limiting layer around a pneumatic pouch. The auxetic skins are manufactured as flexible printed circuit boards with dielectric-laminated electrodes on each auxetic unit cell. By layering multiple sheets and applying a voltage between electrodes on subsequent layers, electroadhesion locks individual auxetic unit cells, modulating the pneumatic pouch's local stiffness and thereby the global output shape. By leveraging ESAs' scalable manufacturing processes, its 10x10 ESA array was found to be the highest degree-of-freedom deformable crust surface as of yet \cite{Wang_Chortos_2022}. Its 20x20 cm$^2$ size enabled whole-hand haptic interaction, although it was only tested to produce simple shapes.

Jang et al. \cite{jang2024dynamically}, shown in Fig. \ref{fig:lit_review_images_hapticsurface}(c.ii), applied SEHAs to build a fingertip-sized deformable crust surface. They patterned a rigid panel with a 5x5 electrode grid for the bottom surface and used a PVC gel sheet embedded with a thin flexible electrode layer for the top haptic surface. The entire display measured 38 x 38 x 1.5 mm$^3$ and weighed only 7 grams. Applying a potential difference between an electrode on the bottom surface and the grounded top electrode causes the PVC gel sheet to compress locally, squeezing a wave of dielectric liquid outwards. Locking every electrode but one squeezes all the liquid into that one cell, generating forces up to 2.4 N and displacements up to 2.0 mm at 1 Hz and 1.0 mm at 30 Hz. Locking groups of cells in sequence could create waves for dynamic haptic experiences and to move light objects around at speeds of up to 140 mm/s.

%% file: 4_adapting_traditional_design_workflows_for_electrostatic_actuators.tex
\section{Design Workflows and Considerations for Electrostatic Actuators} \label{sec:design_workflows}

Designing high-voltage actuators requires careful attention to user safety and comfort. Each HVEA type operates using different voltage regimes, with EOPs using the smallest average RMS voltage across all papers surveyed (287.5 V), followed by ESAs (723 V), DEAs (2978 V), SEHAs (4160 V), and finally EHDPs (7042 V). Supplementary Materials, Sec. S5 lists additional details about this voltage distribution. While higher voltages generally enable greater force outputs, they also introduce additional safety precautions and limit circuit component selection. In this section, we outline typical HVEA design workflows and discuss key considerations for ensuring actuator safety, efficacy, and durability.
% out of the 241 surveyed studies that reported the voltages they used, a little under half (48.5\%, 117/241) used voltages under 1 kV, with 38.5\% (93/241) using voltages under 500 V. 

\subsection{Fabricating HVEAs} \label{subsec:fabricating_hveas}

\revise{HVEAs can be fabricated in many ways, varying from low-fidelity prototyping to professional manufacturing. Accessible and low-cost fabrication pipelines can broaden HVEA adoption within the haptic community. Thus, in this section, we first highlight works that proposed easy-to-implement toolkits or design workflows for each HVEA technology. We then discuss more advanced manufacturing approaches identified in our survey and share practical insights, such as several off-the-shelf starter kits. These resources can help haptics researchers and designers new to HVEAs get started with prototyping.}
% Power supply considerations are omitted here and discussed separately in Sec. \ref{subsubsec:high_voltage_power_supply}.

ESAs are the fastest and most forgiving HVEA to fabricate. Mishra et al. \cite{Mishra_Kumar_Shukla_Parnami_2022} present a low-cost toolkit that required only aluminum foil or copper tape (for the electrodes) and polyimide tape (for the dielectric). They showed a variety of prototypes integrating ESAs into everyday objects, such as a computer mouse and a fish toy, for kinesthetic feedback. 

DEAs also have a simple and customizable layer stack, and Franinović and Franzke \cite{Franinovic_Franzke_2019} provide a helpful toolkit and many examples for integrating carbon black powder (for the electrodes) with VHB foil (for the dielectric). 

\revise{SEHAs require more materials and tools for prototyping, but Mitchell et al. \cite{mitchell2019easy} introduce a fairly low-cost and customizable SEHA fabrication pipeline that adapts an FDM 3D printer to heat-seal thin plastic sheets into a pouch. The pouches can be filled with dielectric liquid (Envirotemp FR3, Cargill) using a syringe, sealed at the fill port with a soldering iron, and coated with carbon black paint to form the electrodes.}

We were not able to find a similar toolkit for EOP EKPs, but Fujii et al. \cite{fujii2021layerpump} present simple instructions for building EHDPs. They use a vinyl cutter to cut copper tape into an interdigitated electrode pattern and using a laser-cut acrylic box to contain the dielectric liquid reservoir.
% , given their novelty within haptics and specialized layer stacks. However
%  and electrode geometries

More advanced manufacturing techniques can further improve scalability, actuator performance, and spatial resolution. Flexible printed circuit boards enable high-resolution, bendable electrode arrays with integrated wiring \cite{Rauf_Bernardo_Follmer_2023, Shen_Rae-Grant_Mullenbach_Harrison_Shultz_2023}, and the circuit layouts can be programmatically generated to easily scale between different array sizes \cite{Rauf_Bernardo_Follmer_2023}. Silicone molding and laser-cut stencils can scalably manufacture customized pouch shapes for SEHAs \cite{Leroy_Shea_2023}. Inspired by micro-electromechanical systems (MEMS) \cite{Rauf_Contreras_Shih_Schindler_Pister_2022, Rauf_Kilberg_Schindler_Park_Pister_2020}, high-spatial-resolution ESAs and DEAs can be fabricated using photolithography \cite{Takagi_Sasaki_Shikida_Sato_2007, Pyo_Ryu_Kyung_Yun_Kwon_2018, Ren_Liu_Lin_Wang_Zhang_2008}, and ultra-thin dielectric layers can be applied onto electrodes using chemical vapor deposition \cite{Shultz_Peshkin_Colgate_2018}. Dielectrics can also be blade cast over large areas by mixing powdered forms of the dielectric with a solvent \cite{Diller_Collins_Majidi_2018, Leroy_Shea_2023, Hinchet_Shea_2020}.
%be used to 
% with controllable thickness 

Several start-ups also sell commercial development kits for HVEAs. ESA touchscreens from Tanvas \cite{tanvastouch} were used in 6 papers in our literature review. ESTAT Actuation originated from a project developing bipolar ESAs for ankle exoskeletons \cite{Diller_Collins_Majidi_2018}, and it sells development kits of their bipolar ESAs in different form factors. Artimus Robotics originated from a project developing SEHA HASELs for soft robotics \cite{acome2018hydraulically}, and they remain active in the research space. They sell SEHA development kits and high voltage power supplies capable of multi-channel HVEA actuation and capacitive sensing. Fluid Reality is in the process of commercializing EOP EKPs \cite{Shen_Rae-Grant_Mullenbach_Harrison_Shultz_2023}.

% Cellulose acetate is a particularly interesting dielectric given its material similarity to paper, and eight papers in our survey took advantage of its flexibility and semi-transparent appearance to fabricate out-of-plane vibrotactile arrays \cite{Bacheva_Firouzeh_Leroy_Balciunaite_Davila_Gabay_Paratore_Bercovici_Shea_Kaigala_2023}.

% HVEAs' versatile manufacturing process also means that every HVEA can be implemented in flexible form factors, including traditionally rigid applications such as ESA touchscreens \cite{Radivojevic_Beecher_Bower_Haque_Andrew_Hasan_Bonaccorso_Ferrari_Henson_2012} and EOP pixel arrays \cite{Yu_Liu_Liu_Lu_Han_Mi_2024}.

\subsection{High Voltage Safety Near the Human Body} \label{subsec:high_voltage_safety}

Similar to conventional electronics, high voltage exposure risk depends mainly on the current intensity-duration and the charge-duration injected into the body. Underwriters Laboratories (UL) and the International Electro-technical Commission (IEC) provide safety standards that are commonly used when verifying the safety of consumer electronics. Both sources conservatively prescribe that 0.5 mA RMS is enough to trigger a startle reaction, 5 mA RMS triggers a ``let-go'' reaction, and 20 mA RMS can trigger ventricular fibrillation. Currents under 20 mA have little to no risk of ventricular fibrillation for any contact duration, and for short durations, hig her currents are also acceptable. For example, both UL and IEC consider 300 mA to have little to no risk of ventricular fibrillation if applied to the skin for less than 25 ms, while 3 A of current is considered ``safe'' if applied for under 100 \textmu s \cite{Reilly_1998}.

High voltages have the potential for greater harm because of their ability to produce high currents and their potential to induce arcing. For back-of-the-envelope calculations, a common approximation, called the human-body model, is that the human body can be estimated as a 1.5 k$\Omega$ resistor and 100 pF capacitor in parallel to ground \cite{mil_std_883_3, jedec_js001_2017}. In addition, while skin can resist brief sparks under 500 V, voltages above 1 kV can induce arcing through desiccated skin \cite{Reilly_1998}.
% , as defined by US MIL-STD-883 and JEDEC JS-001 

Summarizing prior high voltage safety guidelines, we present four design recommendations towards safe human-HVEA interaction: 

(1) The high voltage power supply’s maximum output current should ideally be under 20 mA. If larger currents are needed to ensure device performance, additional safety circuitry should be implemented to shut off the current within a time specified by the UL and IEC guidelines (in general, within a few milliseconds) \cite{Reilly_1998}. 

(2) The HVEA’s capacitance should be kept below a critical limit to avoid spontaneous discharge into the human body if accidentally touched. Assuming users have dry skin, Pourazadi et al. \cite{Pourazadi_Shagerdmootaab_Chan_Moallem_Menon_2017} conservatively found this limit to be 1.85 \textmu F for HVEAs charged at 1 kV and 80 nF at 10 kV. For context, 80 nF is equivalent to a DEA with a 250 \textmu m thick VHB 4905 dielectric and a 0.7 m$^2$ cross-sectional area; it is uncommon to reach these limits without stacking many HVEAs in series. We provide equations for each HVEA's capacitance in the Supplementary Materials, Sec. S1 and more details about these capacitance limits in the Supplementary Materials, Sec. S7. 

(3) High voltage electrodes and wires should have a layer of insulation (ideally two, in case of catastrophic dielectric failure), each rated for at least 2$\times$ the nominal operating voltage \cite{IEC62368_1_2018}. Where possible, the grounded electrode should be oriented closer to the user. For example, polyimide has a rated dielectric strength of about 100 V/\textmu m, so a 1 kV application should use 1-2 layers of 20 \textmu m (0.8 mil) thick polyimide tape to insulate the electrodes. 
% IEC 60502 \cite{IEC_60502_2} recommends that insulation thickness should be set to withstand voltage spikes at least 3-4$\times$ the nominal voltage. % 15.3.4

% U.2.1 Electric strength, IEC 60950-1

(4) Researchers should follow their institution's safety regulations for working with high voltages. The IEEE Std 510-1983 \cite{ieee_510_1983} provides general guidance, such as enclosing high voltage setups within a box or barrier, posting ``DANGER - HIGH VOLTAGE'' signs, discharging all capacitive devices after use, and wearing rubber gloves when handling live wires.
% These limits increase to 37.1 \textmu F at 1 kV and 0.258 \textmu F at 10 kV if we assume the HVEAs can only be touched after the 20 mA-rated high voltage power supply is disconnected.
% , which provides insulation guidelines for power cables carrying 1-30 kV,

We refer interested readers to the excellent reviews by Pourazadi et al. \cite{Pourazadi_Shagerdmootaab_Chan_Moallem_Menon_2017} and Reilly \cite{Reilly_1998} on specific design guidance before testing any devices near the upper threshold of these safety limitations on humans.

% \ran{Although HVEAs generally operate at high voltages, raising potential concerns for applying to the human body, they can be effectively insulated with polymers, elastomers, or flexible electronics to satisfy standard safety regulations like CE, UL, and IEC (We have the white paper from Artimus but not sure how sufficient it is \footnote{https://www.artimusrobotics.com/\_files/ugd/c39046\_c67565cc62ca456e9674adbfb02d17b3.pdf}} 

\subsection{High Voltage Circuitry Component Selection} \label{subsec:high_voltage_circuitry_component_selection}
% While high voltage actuators have become increasingly rare in hobbyist electronics, due largely to the decreasing size and threshold voltages of transistors, they remain common in certain industrial applications. Kenotrons, or thermionic vacuum rectifying hot-cathode diodes, were \href{https://medicine.lau.edu.lb/related-entities/zahi-hakim-museum/files/Kenotrons.pdf}{used until the 1970s in the high tension circuits of x-ray tubes}. Cathode ray tube televisions use potential differences in the thousands of volts to accelerate electrodes, which then are angled by magnetic fields to scan an image on the television screen. Night vision goggles work using a similar principle, using a potential difference of about 5000 V to amplify the signal from incoming photons up to threshold visible to the naked eye. Digital micromirror devices in many modern projectors use arrays of hundreds of micro-electromechanical system (MEMS) planar electrostatic actuators to individually tilt tiny silicon mirrors, forming an output projected image \cite{Patterson_Hah_Fujino_Piyawattanametha_Wu_2004}. Inkjet printers control the movement of electrically charged ink droplets using the electric field within capacitor plates \href{http://spiff.rit.edu/classes/phys213/lectures/inkjet/inkjet_long.html}{held at several hundred volts}. The latter two applications, in particular, have led to the develop of mass-market, multi-channel high voltage driver chips powered by High-Voltage Complementary Metal–Oxide–Semiconductors (HVCMOS).

\subsubsection{High Voltage Power Supply} \label{subsubsec:high_voltage_power_supply}
There are two primary methods to output a high-voltage drive signal for HVEAs. If the application only requires a constant voltage for on/off haptic signals, a DC-DC converter is sufficient. Low-cost, low-power DC-DC converters can be implemented manually using Cockcroft-Walton multipliers \cite{Iriarte_Ezcurdia_Elizondo_Irisarri_Hemmerling_Ortiz_Marzo_2024, Mujibiya_2015} or flyback transformers \cite{Ji_Liu_Cacucciolo_Civet_ElHaitami_Cantin_Perriard_Shea_2021}, and higher-power DC-DC converters can be purchased commercially. Commercial modules can be incredibly compact, only a few cm$^3$ in volume, and they can be driven by batteries for ultra-portable and wearable applications.
%  for a few tens to hundreds of dollars USD.

\revision{Variable-intensity haptic sensations}, such as sinusoidal vibration patterns, are typically generated by amplifying a scaled-down version of the signal from a microcontroller or function generator up to the desired high voltage range. These amplifiers are often expensive and bulky, with most commercial solutions coming as rack-mounted or floor tower units. However, open source projects, such as the Peta-pico-Voltron \cite{Schlatter_Illenberger_Rosset_2018}, and commercial UltraVolt amplifiers from Advanced Energy have been used in prior work for portable applications. Supplementary Materials, Sec. S5 summarizes our survey's most common high voltage power supplies and amplifiers, including several portable and relatively affordable options. 
% If the application requires variable-intensity haptic sensations, such as sinusoidal vibration patterns, the typical solution is to amplify a scaled-down version of the signal from a microcontroller or function generator up to the desired high voltage range. These amplifiers tend to be more expensive and bulky, with most commercial solutions coming as rack-mounted or floor tower units. However, open source projects such as the Peta-pico-Voltron \cite{Schlatter_Illenberger_Rosset_2018} and commercial amplifiers from Advanced Energy's UltraVolt line have been deployed in prior work for portable applications. Supplementary Materials, Sec. S5 summarizes the most common high voltage power supplies and amplifiers used in the literature, including several portable and relatively affordable options. 

Most HVEAs operate using wall power or batteries. However, Qu et al. \cite{Qu_Ma_Shi_Li_Zheng_Wang_Liu_Fan_Chen_Li_2021} present an interesting alternative, powering a DEA dot array using a triboelectric nanogenerator (TENG) built from a 16 cm$^2$ polyimide sheet on top of aluminum foil and an acrylic block. The TENG generated voltages up to 3.25 kV and currents up to 2 \textmu A, and the dot array could be refreshed every time a user pressed on the TENG. They refreshed the dot array 20,000 times and found that the TENG and the dot array showed no performance degradation. 

\subsubsection{High-Voltage Logic Circuitry} \label{subsubsec:high_voltage_control_circuitry}
Many applications pair a high-voltage power supply with additional logic circuitry to quickly charge and discharge HVEAs. This is required because many high-voltage power supplies require a small start-up time before their output voltage settles. In addition, compact DC-DC converters are often worse at sinking current than sourcing it, and without a separate discharge circuit, HVEA dielectrics can take seconds to minutes to fully reverse their space charge polarization after the high voltage is removed. HVEAs that accumulate space charge over many actuation cycles often worsen in performance until the space charge is dissipated \cite{Diller_Collins_Majidi_2018, Johnson_Naris_Sundaram_Volchko_Ly_Mitchell_Acome_Kellaris_Keplinger_Correll_2023}, so for many applications it can often be useful to have additional circuitry to quickly discharge an HVEA.
% , which in turn means that the HVEA takes longer to actuate

Common logic circuits used in prior literature include common source amplifiers, push–pull output stages such as half bridges and full bridges, and optocoupler relays. We include a comparison of the pros and cons of different logic circuits in terms of implementation complexity, rise and fall time, and cost in the Supplementary Materials, Sec. S6. In summary, half bridges comprising an NMOS and PMOS transistor work well up to about 600 V, half bridges comprising three NMOS transistors work well up to about 4 kV, and optocoupler relays are best for voltages above 4 kV. Full bridges allow for faster discharging than half bridges, especially for power supplies that are worse at sinking currents than sourcing them, but they require twice as many transistors and twice the board space.
% circuit diagrams and 

\subsection{Self-Sensing Circuitry} \label{subsec:selfsensing_circuitry}
Every HVEA can be modeled as a capacitor in series with its electrodes' resistance and in parallel with its dielectric's resistance. Although this paper primarily focuses on HVEAs' use as actuators, an HVEA's capacitance and resistance can be related to its internal deformation for proprioceptive sensing and, depending on the application, to the proximity of nearby surfaces for exteroceptive sensing. The Supplementary Materials, Sec. S1 provides formulas to relate the measured capacitance and resistance to HVEA geometry. Prior methods for simultaneous high voltage actuation and sensing of HVEAs can be grouped into three main categories: resistive sensing, capacitive sensing, and complex impedance sensing. 

Resistive sensing is typically implemented by dividing the high voltage power supply's output voltage and current waveforms, where the current can be measured from the voltage drop over a small resistor in series with the HVEA. Alternatively, a resistive divider can be formed between the HVEA and an external resistor, and the voltage across the external resistor can be used to estimate the HVEA's resistance \cite{Nakamura_Yamamoto_2016, Matysek_Haus_Moessinger_Brokken_Lotz_Schlaak_2011}. Resistive sensing methods are fast and reliable in rigid applications such as touchscreens. However, the resistance of soft electrodes often degrades over time or after large strain, limiting resistive sensing's reliability over long durations.
% , and burden voltage over the current measurement resistor can cause measurement inaccuracy.

Capacitive sensing methods work by adding a low-voltage, high-frequency AC waveform to the high-voltage actuation signal. The capacitance can be measured from the voltage drop over the HVEA through a capacitive divider \cite{Trase_Xu_Chen_Tan_Zhang_2020, Trase_Tan_Chen_Zhang_2021, Matysek_Lotz_Winterstein_Schlaak_2009}, \revision{from the AC input's phase shift through the HVEA} \cite{acome2018hydraulically, Haga_Sugimoto_Yang_Sasaki_Asai_Shigemura_2019}, or by comparing the voltage rise and fall time constants through two different load resistors \cite{Gratz-Kelly_Kruger_Rizzello_Seelecke_Moretti_2023}. Capacitance sensing is generally more reliable than resistance sensing for macro-scale HVEAs. However, it can be noisy because of capacitive coupling with the environment, and many algorithms still assume that the electrode resistance is constant and does not degrade over time.

The most robust (and computationally intensive) methods measure an HVEA's complex impedance, including both its resistance and capacitance. These methods typically sample the voltage and current output of the high-voltage power supply while also adding a low-voltage, high-frequency sinusoidal sensing signal to the high-voltage actuation signal. A recursive least squares method can be implemented to extract the HVEA's resistance and capacitance in real time for generic high voltage drive signals \cite{Gratz-Kelly_Kruger_Seelecke_Rizzello_Moretti_2024,Gratz-Kelly_Kruger_Seelecke_Rizzello_Moretti_2023}. Linearized methods can further improve robustness if the high voltage drive signal is known to be constant (DC), linear, or sinusoidal \cite{Chen_Besier_Anderson_McKay_2014, Knoop_Rossiter_Assaf_2015}.

Very few papers in our literature review have extended these single-channel implementations of simultaneous actuation and sensing to larger HVEA arrays. \revision{The largest array with simultaneous actuation and sensing in our literature review comprised 32 ESAs \cite{Rauf_2025}, and the next-largest comprised 6 SEHAs \cite{HASEL_Actuators}.} Many papers either remove the high voltage signal while sensing, as done by Haga et al. \cite{Haga_Sugimoto_Yang_Sasaki_Asai_Shigemura_2019} across a 42 x 32 electrode array in an ESA touchscreen, or use alternative sensing methods such as magnetic \cite{Johnson_Naris_Sundaram_Volchko_Ly_Mitchell_Acome_Kellaris_Keplinger_Correll_2023} or resistive strain sensors \cite{Fishman_Catsis_Homer_Rossiter_2018}. Future work should investigate more scalable circuitry and calibration techniques for multi-channel simultaneous actuation and sensing.
% The largest array with simultaneous actuation and sensing in our literature review only comprised 6 SEHAs \cite{HASEL_Actuators}.

% 4 DEAs \cite{Trase_Tan_Chen_Zhang_2021}

%% file: 5_conclusions_and_future_directions.tex
\section{Conclusions and Outlook}

 % Comments from Teng: : i kind of feel that there should be some summarization before future direction, and it shall recall to the Intro's "design, ergonomics, and perceptual challenges". Do we solve them or have answered to them? maybe highlight the importance of learning this survey for hci people doing haptics and its practical value.

The field of high-voltage electrostatic actuators has greatly expanded during the past decade, showcasing novel technologies, diverse applications, and interdisciplinary collaboration. Through this systematic literature review, we synthesized how these actuators have been designed, fabricated, and integrated into haptic interfaces, while identifying trends in form factors, applications, and safety considerations. 
% the form factors, applications, safety guidelines, and design approaches explored by prior work that integrates HVEAs into haptic user interfaces.

Our review reveals that HVEAs offer distinctive advantages over conventional haptic actuators. By classifying prior work by touch modality and further mapping it across HVEA form factors, technologies, and application domains, our analysis highlights how HVEAs address long-standing design and ergonomic challenges in haptics. Their thin, compact, and compliant structures allow them to conform naturally to the human body and integrate seamlessly into the surrounding environment. HVEAs' ability to render versatile sensations across multiple haptic modalities opens up novel perceptual possibilities, ranging from kinesthetic feedback and controllable normal or shear force to high-bandwidth, yet near-silent vibrations. Their energy-efficient operation makes them ideal for untethered use across wearables and handheld devices. 

HVEAs can also be easily manufactured at scale, and a number of studies in our review manufactured arrays of hundreds to thousands of individually addressable actuators using monolithic manufacturing techniques. For HCI and haptics researchers, understanding these capabilities is not only technically valuable but practically empowering, enabling the design and deployment of more ergonomic, expressive, and energy-efficient haptic systems across diverse applications.
%in a single production run
% domains

% Their compact, compliant, and customizable forms are readily adaptable to a wide range of on-body and world-grounded contexts. Their quiet and energy-efficient operation makes them advantageous for untethered use cases such as braille displays, tablet touchscreens, and wearable gloves. Their high bandwidth, often in the tens to hundreds of Hertz, allows them to readily transition between low-frequency and vibrotactile cutaneous feedback to produce a wide range of haptic interactions. HVEAs can also be easily manufactured at scale, and a number of studies in our review manufactured arrays of hundreds to thousands of individually addressable actuators in a single production run using monolithic manufacturing techniques such as laser cutting, printed circuit board manufacturing, and low-pressure chemical vapor deposition (LPCVD).

Despite their potential, several major barriers still exist to widespread HVEA deployment. Their specialized manufacturing techniques and drive circuitry have limited HVEA's commercial availability and increased the barrier to entry for haptics researchers. HVEAs have primarily planar output geometries, and prior literature has largely explored their ability to produce cutaneous feedback rather than kinesthetic feedback. Finally, the field continues to suffer from a lack of interdisciplinary knowledge exchange between communities such as materials science, HCI, and haptics.

\subsection{Future Directions}

Building on the takeaways from this survey, we identify several future directions for advancing the integration of HVEAs in haptic interfaces and applications. These perspectives are informed by our author team's specialized expertise in developing HVEA and haptic systems, as well as by collective insights from the HCI community. In particular, we organized a workshop at the 2023 ACM Symposium on User Interface Software and Technology (UIST) \cite{Leithinger_Zhou_Acome_Rauf_Han_Shultz_Mullenbach_2023}, which brought together 20 participants from academia and industry over a day of interactive demos, focused discussions, and hands-on ideation, where we identified key challenges and opportunities of HVEAs for haptics. In the following, we outline these directions, aiming to pave pathways for future exploration in HVEA-based haptic research.

\subsubsection{Fostering Interdisciplinary Collaboration and Knowledge Synthesis}

Advancing HVEA technologies for haptic applications requires close interdisciplinary collaboration between communities, including but not limited to material science, mechanical engineering, physics, HCI, haptics, and design. Such knowledge synthesis can inform HCI and haptic researchers about cutting-edge advancements, such as new high-$\kappa$ dielectric materials and mechanisms for lower-voltage operation \cite{gravert2024low}. Ionic electro-active polymers, such as conductive polymer networks \cite{Wang_Sotzing_Lee_Chortos_2023}, ionic polymer-metal composites \cite{Feng_Hou_2018}, and stimuli-responsive hydrogels \cite{Paschew_Richter_2010}, are emerging actuators that can leverage HVEA's flexible form factors and integrated self-sensing at voltages under 10 V, and although manufacturing these at scale remains challenging they remain a promising area for future research. Deeper interdisciplinary collaboration also allows HCI researchers and designers to communicate their goals, priorities, considerations, and concerns to material and mechanical experts, accelerating the translation of HVEAs into real-world applications.
% . This exchange can help align fundamental research on material properties with practical haptic design needs, thereby 

Collaboration within the HVEA domain itself also holds great potential. Although all HVEAs share similar electrostatic actuation principles and high-voltage operation, most of the devices in our survey employ only a single HVEA technology. By leveraging the customizability of these actuators, particularly their thin, flexible, and layer-based architectures, future systems could combine multiple HVEA types to create multimodal haptic devices. For example, while kinesthetic feedback has so far been explored primarily through ESAs, integrating SEHAs or DEAs could enable richer, composite tactile and kinesthetic experiences.

% haptic/HCI researchers shall collaborate closely with materials scientists. First, we need to stay informed about cutting-edge advancements in the field, such as HVEAs, which now operate at significantly lower voltages due to improved dielectric materials. Second, we must clearly communicate our goals, priorities, and concerns to materials experts. This interdisciplinary collaboration will drive HVEAs toward practical, real-world applications.

% For future directions, I tend to think more engineering is needed. Specifically high d constant materials are critical (an what enabled low voltage EOPs). It usually comes down to how much energy you can store in the E-field, and instead of raising the voltage, you can always raise the d-constant. 

% I feel one very strong future direction is leveraging the customizability of these actuators to compose them into multi-model haptic devices. Especially because they are thin and flexible, so that can have several layers to work together.

% We are excited about the potential of HVEAs and how they will transform interaction with computing systems. We see the identification of these challenges as a positive step in focusing research efforts towards this common goal. These challenges will, no doubt, evolve as technology matures and our understanding of users increases. We look forward to future interdisciplinary collaborations continuing to advance this research going forward.

\revise{
\subsubsection{Developing Accessible Prototyping and Haptic Design Toolkits}

In our workshop \cite{Leithinger_Zhou_Acome_Rauf_Han_Shultz_Mullenbach_2023}, a recurring theme was that haptics researchers often consider HVEAs to have a high barrier to entry. This is due to a number of factors, such as the lack of off-the-shelf commercial HVEAs for fast prototyping, the mixed electromechanical experience required to fully understand how HVEAs operate, and the intimidation of working with high voltages. Although we attempt to address some of these fabrication workflows in Sec. \ref{sec:design_workflows}, future work should explore how to best develop and sustain accessible HVEA prototyping toolkits for HCI and haptics researchers. 

Beyond actuator prototyping, there is also a need for design-oriented toolkits that support the creative exploration of HVEA-based haptics. A key strength of HVEAs lies in their capability to provide versatile and expressive touch sensations. To fully harness this potential, haptic researchers and designers must be able to experience and design with HVEA-delivered cues directly, without needing the complex technical knowledge and skills to build them from scratch. Such toolkits could follow the model of open-source projects like FlowIO \cite{Shtarbanov_2021}, TactorBots \cite{zhou2023tactorbots}, and ANISMA \cite{messerschimidt2022anisma}, which freely loaned development and design kits to academics in return for hardware, software, and tutorial contributions that could last beyond the project's original developer. 
}

%Such toolkits could follow the model of open-source projects like FlowIO \cite{Shtarbanov_2021}, which freely loaned development kits to academics in return for hardware, software, and tutorial contributions that could last beyond the project's original developer.
\revise{
\subsubsection{Characterizing the User Experience and Felt Quality of HVEA-Delivered Haptics}
While HVEA haptic innovations have achieved remarkable technical progress, their evaluation of haptic performances has largely centered on engineering matrices such as bandwidth, displacement, blocking force, or response rate. Their user studies typically emphasize controlled measures of perceivability, pattern discrimination, and XR simulation's realism. Beyond these standardized metrics, however, many papers report that HVEAs offer a user experience and felt quality unique from traditional actuators.
% While these methods are standard for assessing haptic interfaces, a deeper characterization of the user experience and felt quality of HVEA-delivered haptics is essential for establishing their distinct value. 
% With such understanding, HVEAs could advance not only through optimal mechanical performance but also by revealing their unique tactile richness and expressive potential. 

% In Fukushima and Kajimoto's study on piloerection \cite{Fukushima_Kajimoto_2012}, for example, they found participants often explained their experience through metaphors
\revision{Several} studies have begun to investigate how HVEA-delivered sensations are perceived and described by users. In Fukushima and Kajimoto's study on piloerection \cite{Fukushima_Kajimoto_2012}, for example, they found participants often explained their experience through metaphors, describing the artificial goosebumps as ``a wind sensation'' or that ``I felt like I was wearing aura.'' Using electrostatic frictional displays, Tomita et al. \cite{Tomita_Saga_Kajimoto_Vasilache_Takahashi_2018} had Japanese users describe different friction patterns and frequencies using onomonopoeias such as ``gori-gori'' (dry and crunchy) or ``tsuru-tsuru'' (smooth and slippery). Their findings suggest that more intuitive, symbolic language can allow users to better convey how they feel about tactile sensations. Future research can learn from these early insights to better convey the experiential quality of HVEA-based haptics. Doing so would make their features and benefits more compelling to the broader haptics community, inviting and inspiring further exploration of this emerging class of technologies.}

\revise{
\subsubsection{Enhancing Wearability by Leveraging HVEA Properties}

HVEA actuators hold great potential for wearable haptic devices, as they are thin, soft, compliant, and self-contained. Their customizable scale and geometry make them highly adaptable, not only as direct replacements for conventional actuators within rigid enclosures and housings \cite{sanchez2024cutaneous}, but also as components embedded in soft materials such as silicone rubber gloves \cite{song2019pneumatic, Hinchet_Shea_2022}. Several works have demonstrated direct skin attachment of thin-film HVEA actuators using adhesives \cite{Leroy_Shea_2023, Youn_Jang_Hwang_Pei_Yun_Kyung_2025}, highlighting their conformability for on-skin haptics.

Woven textiles represent another promising substrate for haptic garments and wearables, and many HVEAs can be readily adapted to fiber-like or sheet-based form factors. ESAs have been integrated into textiles either using conductive fabric \cite{Kitagishi_Hiraki_Nakamura_Ishiguro_Rekimoto_2024} or by adhering dielectric films to textiles using conductive tape \cite{Hinchet_Shea_2020}, while Smith et al. \cite{Smith_Cacucciolo_Shea_2023} demonstrated woven EHDP fibers directly into a textile for distributed thermal feedback. These examples illustrate strong compatibility between HVEAs and textile-based haptic wearables. However, though Hinchet and Shea \cite{Hinchet_Shea_2020} showed that their textile electrostatic clutch could be washed in soapy water to recover its adhesive performance after wear, few studies have systematically examined the washability and long-term durability of HVEA textile systems. Future work should build upon those foundations, enabling wearable haptic devices that are more ergonomic, robust, and long-lasting for everyday use. 
}
%  while further enhancing the wearability of HVEA by leveraging its special properties

% De Rossi et al. \cite{Rossi_Carpi_Lorussi_Scilingo_Tognetti_2009} presented a concept of kinesthetic garments using rolled DEAs as woven fibers, but they only tested the rolled DEAs individually and we found no evidence of follow-up work since the paper's publication in 2009. Despite these proof-of-concept examples, little research has been conducted into improving the manufacturing scalability and durability of HVEA actuated fibers.

% (washing textiles in soapy water, HASEL = self-healing but can be punctured)

% \subsubsection{Studying Durability and Reliability in Practical Contexts}

% % (washing textiles in soapy water, HASEL = self-healing but can be punctured)

\revise{
\subsubsection{Integrating HVEAs into Real-World Applications}
While fundamental research is crucial for maturing HVEA technologies, there is also a pressing need for application-focused studies that demonstrate their benefits in real-world contexts. Beyond measuring and quantifying technical capabilities, future work should investigate how these advantages translate into improved user experiences and enhanced practical outcomes. 
% Such exploration can not only advance the existing haptic applications but also potentially open up new forms of interactions and use cases that are not possible with conventional actuators. 

In the realm of accessibility, HVEAs have already shown promise in building more compact and scalable refreshable braille displays \cite{Ren_Liu_Lin_Wang_Zhang_2008} and towards providing more salient navigation cues for people with low vision \cite{Feitl_Kreimeier_Gotzelmann_2022, Bateman_Zhao_Bajcsy_Jennings_Toth_Cohen_Horton_Khattar_Kuo_Lee_2018} In medical contexts, several studies have leveraged HVEAs’ compatibility with magnetic resonance environments and their suitability for compact exoskeletons \cite{yamamoto2005evaluation, Han_Bae_Gregoriou_Ploch_Goldman_Glover_Daniel_Cutkosky_2018, Mahmoud_Halabi_Ahmed_Sofela_Naumov_2021}, pointing toward further opportunities in rehabilitation and healthcare technologies. As haptic interfaces are increasingly explored for enhancing well-being, such as stress relief and emotion regulation, HVEAs offer advantages over commonly used vibration motors through their more silent, nuanced, and expressive tactile feedback. 
% . Future work could extend this potential to other assistive applications, such as

Teleoperation of humanoid robots represents another emerging frontier for HVEAs. Their high force density, responsiveness, and low latency make them ideal for gloves or suits that deliver instantaneous, realistic feedback from remote environments. Through interdisciplinary collaboration, particularly with HCI researchers and designers, we are confident and looking forward to seeing how future explorations can uncover novel application scenarios and broaden the design space, enabling HVEA-based haptic technologies to meaningfully enhance both everyday and domain-specific experiences.
}

\revision{Future work should further characterize the durability and robustness of HVEAs in this real-world environments. ESAs, DEAs, and SEHAs have already demonstrated lifetimes in the millions of cycles in controlled lab settings \cite{Diller_Collins_Majidi_2018, acome2018hydraulically, Tang_Du_Jiang_Wang_Liu_Zhao_2024}. However, sweat buildup and environmental factors such as humidity and temperature can affect HVEA performance, particularly in wearable applications \cite{AliAbbasi_Muzammil_Sirin_Lefevre_Martinsen_Basdogan_2024}, and their impact should be better understood before practical use.}